\documentclass[10pt,twocolumn,letterpaper]{article}

\usepackage{cvpr}              % To produce the CAMERA-READY version
\usepackage[accsupp]{axessibility}  % Improves PDF readability for those with disabilities.
\usepackage{wrapfig}
\usepackage{caption}
\usepackage{graphicx}
\usepackage{makecell}
\usepackage{amsthm}
\usepackage{amssymb}
\usepackage{pifont}
\usepackage{bm}
\usepackage{colortbl}
\usepackage{amsmath}
\usepackage{multirow}
\usepackage{enumitem}
\usepackage{bbding}
\usepackage{color}
\usepackage{natbib}
\usepackage{algorithm}
\usepackage{listings}
\usepackage{booktabs}   % \toprule \midrule \bottomrule \cmidrule
\usepackage{array}
\usepackage[normalem]{ulem}
\usepackage[table]{xcolor}
\usepackage{multirow}

\def\modelname{\texttt{SegCompass}}
\usepackage[table]{xcolor}
\definecolor{catyellow}{RGB}{255,248,220}

\makeatletter
\newcommand{\minihead}[1]{%
  \if@nobreak\else
    \par\addvspace{0.10\baselineskip}% space between paragraphs
  \fi
  \noindent\textbf{#1}\;\,%
}
\makeatother

\usepackage{wrapfig}
\newcommand{\SetEqBlank}{
}
\newcommand{\Lgrpo}{\mathcal{L}_{\text{\scriptsize \textsc{grpo}}}}
\newcommand{\Lppo}{\mathcal{L}_{\text{\scriptsize \textsc{ppo}}}}
\newcommand{\Lseg}{\mathcal{L}_{\text{\scriptsize \textsc{seg}}}}
\newcommand{\Lsae}{\mathcal{L}_{\text{\scriptsize \textsc{sae}}}}
\newcommand{\Lconf}{\mathcal{L}_{\text{\scriptsize \textsc{conf}}}}

\newcommand{\ximg}{\bm{x}_{\text{img}}}
\newcommand{\xtxt}{\bm{x}_{\text{txt}}}
\newcommand{\Mgt}{\bm{M}_{\text{gt}}}
\newcommand{\yoneT}{\bm{y}_{\scriptscriptstyle 1:T}}
\newcommand{\policy}{\pi_{\bm{\theta}}}
\newcommand{\yi}{\bm{y}_{\scriptscriptstyle 1:T_i}^{(i)}}
\newcommand{\heatmap}{ \left( \mathcal{H} _k \right) _{k=1}^{K_s}}
\newcommand{\confidence}{ \left(c_k \right) _{k=1}^{K_s}}
\newcommand{\confidenceGT}{ \left(y_k \right) _{k=1}^{K_s}}
\newcommand{\heatconf}{ \left(\mathcal{H}_k, \, c_k \right) _{k=1}^{K_s}}
\newcommand{\Fdec}{\mathcal{F}_{\text{dec}}}
\newcommand{\Fenc}{\mathcal{F}_{\text{enc}}}
\newcommand{\Fslot}{\mathcal{F}_{\text{slot}}}
\newcommand{\ConcenEmbed}{ \left( \bm{e}_k \right) _{k=1}^{K_s}}
\newcommand{\ConceptRepre}{ \left(\bm{r}_k \right) _{k=1}^{K_s}}
\newcommand{\Esae}{\mathcal{E}_{\text{sae}}}
\newcommand{\Dsae}{\mathcal{D}_{\text{sae}}}
\newcommand{\SuppZ}{\mathcal{S}( \bm{z})}

\definecolor{cvprblue}{rgb}{0.21,0.49,0.74}
\usepackage[pagebackref,breaklinks,colorlinks,allcolors=cvprblue]{hyperref}

\def\paperID{} % *** Enter the Paper ID here
\def\confName{CVPR}
\def\confYear{2026}

\title{\modelname{}: Exploring Interpretable Alignment \\with Sparse Autoencoders for Enhanced Reasoning Segmentation}

\author{Zhenyu Lu$^{1,2,5}$, 
Liupeng Li$^{3,2}$, 
Jinpeng Wang$^{3,}\thanks{Jinpeng Wang and Yaowei Wang are corresponding authors.}\;$, 
Haoqian Kang$^{6,3}\thanks{Work done during the undergraduate internship at HITSZ.}\;$,
Yan Feng$^{4}$, 
Ke Chen$^{2}$, 
Yaowei Wang$^{3,2,*}$ \\
{\normalsize ${}^1$Shenzhen Institutes of Advanced Technology, Chinese Academy of Sciences} \qquad
{\normalsize ${}^2$Peng Cheng Laboratory}\\
{\normalsize ${}^3$Harbin Institute of Technology, Shenzhen}\qquad
{\normalsize ${}^4$Meituan, Beijing}\qquad
{\normalsize ${}^5$University of Chinese Academy of Sciences}\\
{\normalsize ${}^6$College of Computer Science and Technology, Jilin University}\\
{\normalsize \texttt{zhenyulu22@m.fudan.edu.cn; wangjp26@gmail.com; wangyaowei@hit.edu.cn}}
}

\begin{document}
\maketitle

\begin{abstract}
While large language models provide strong compositional reasoning, existing reasoning segmentation pipelines fail to transparently connect this reasoning to visual perception.
Current methods, such as \emph{latent query alignment}, are end-to-end yet opaque ``black boxes".
Conversely, \emph{textual localization readout} is merely readable, not truly interpretable, often functioning as an unconstrained post-hoc step.
To bridge this interpretability gap, we propose \textbf{\modelname{}}, an end-to-end model that leverages a \textbf{Sparse Autoencoder (SAE)} to forge an explicit, interpretable, and differentiable alignment pathway.
Given an image-instruction pair, \modelname{} first generates a chain-of-thought (CoT) trace.
The core of our method is an SAE that maps both the CoT and visual tokens into a shared, high-dimensional sparse concept space.
A query codebook selects salient concepts from this space, which are then spatially grounded by a slot mapper into a multi-slot heatmap that guides the final mask decoder.
The entire model is trained jointly, unifying reinforcement learning for the reasoning path with standard segmentation supervision.
This SAE-driven interface provides a ``white-box" connection that is significantly more traceable than latent queries and more coherent than textual readouts.
Extensive experiments on five challenging benchmarks demonstrate that \modelname{} matches or surpasses state-of-the-art performance.
Crucially, our visual and quantitative analyses show a strong correlation between the quality of the learned sparse concepts and final mask accuracy, confirming that \modelname{} achieves superior results \emph{through} its enhanced and inspectable alignment.
% Faithful code will be released publicly.
Code is available at \texttt{\url{https://github.com/ZhenyuLU-Heliodore/SegCompass}}.
\end{abstract}

\begin{figure*}[t]
\captionsetup{aboveskip=5pt, belowskip=-12pt}
    \centering
    \includegraphics[width=\textwidth]{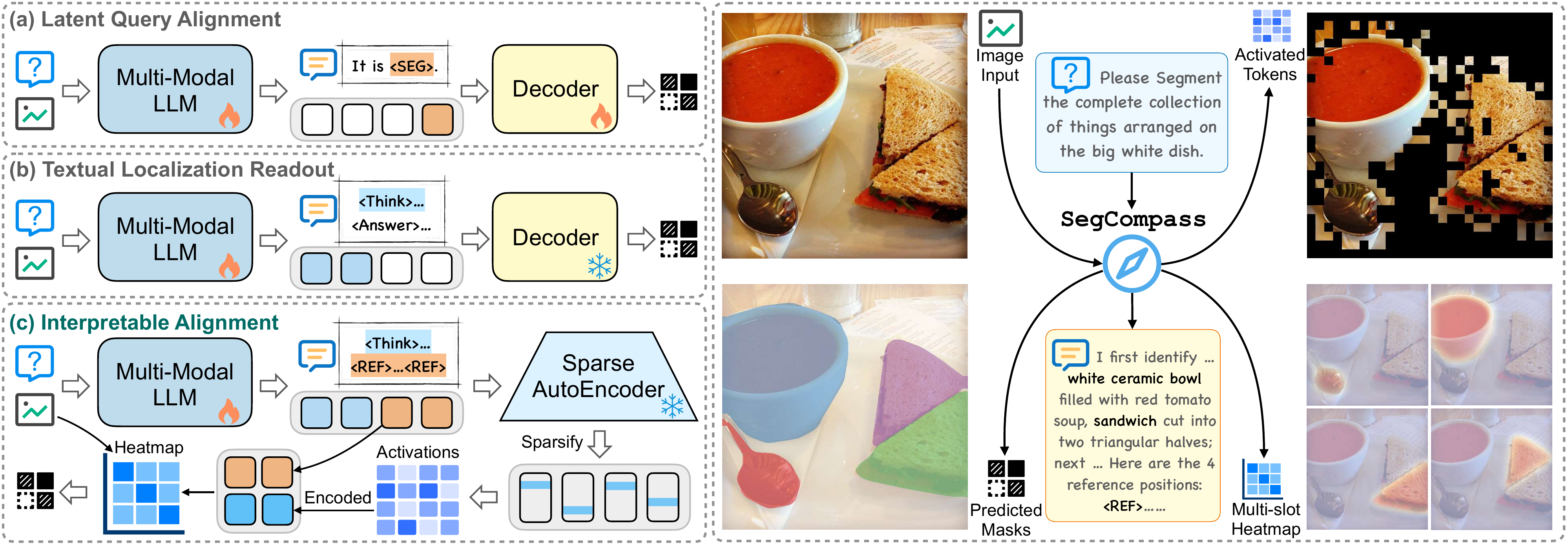}
\caption{
\textbf{\underline{Left}:} Comparison of alignment strategies for reasoning segmentation.
(a) \textbf{Latent Query Alignment} relies on opaque latent vectors.
(b) \textbf{Textual Localization Readout} uses unconstrained textual cues as a post-hoc signal.
(c) Our proposed \textbf{Interpretable Alignment} introduces a SAE as an interpretable bridge, which transforms reasoning tokens into sparse concept activations. These activations generate a multi-slot heatmap, explicitly linking the model's reasoning process to the final segmentation mask.
\textbf{\underline{Right}:} \modelname{}'s interpretable alignment interface. 
The core of our method is visibly linking this reasoning to perception: the concepts identified in the CoT (e.g., ``white ceramic bowl") are explicitly grounded onto the image via a Multi-slot Heatmap.
This provides a white-box view of how the model comprehends and localizes each instruction part, with intermediate Activated Tokens visualizing the sparse concept space.
}
    \label{fig:teaser}
\end{figure*}

\section{Introduction}
\label{sec:introduction}

Reasoning segmentation~\citep{lisa_2024, segmentation_survey} is a foundational component of vision systems guided by natural language instructions.
Beyond simple captioning, many real-world scenarios, such as robotic manipulation and video analysis, necessitate spatial localization based on complex, multi-step instructions (e.g., “Segment the mug that matches the plate’s color and is closest to the sink”).
While recent advances in large language models~\citep{qwenvl_2023, qwen2.5_vl, llava_2023} exhibit powerful compositional and commonsense reasoning, this capability remains inadequately integrated with dense visual perception in current pipelines.
Consequently, the interpretability and controllability of \emph{how the model infers and locates} targets have become critical yet challenging research aspects.

Current work on reasoning segmentation typically follows one of two routes for connecting reasoning outputs with the segmentation module.
\emph{\textbf{Latent query alignment}} approaches~\citep{perceptiongpt_2024, lisa_2024, LIRA, RAS_2025} map hidden representations from the reasoning module to latent queries that interact with visual features to predict masks.
While this route is trained end-to-end, it remains opaque: intermediate decisions are encapsulated and function as a black box.
\emph{\textbf{Textual localization readout}} approaches~\citep{sam4mllm_2024, text4seg_2025, segzero_2025, VisionReasoner} first generate discrete localization tokens (e.g., box coordinates or patch indices), often via chain-of-thought (CoT)~\citep{deepseek-r1_2025, MCoT_Survey_2025} rollouts. These tokens are then utilized in what is effectively an \emph{independent post-processing step} to prompt the segmentation module.
The rationale in this second route is merely readable, yet not truly interpretable: the CoT is unconstrained, the derivation of spatial cues is obscure, and the textual tokens themselves often carry insufficient semantic detail.
In short, a mechanism that provides an interpretable and verifiable connection from the reasoning process to the final mask prediction remains elusive.

Motivated by this gap, we seek a mechanism that can create a "white-box" connection between reasoning and perception. We believe that the discrete, high-dimensional, and semantically meaningful features learned by \emph{Sparse Autoencoders} (SAEs)~\citep{sae_2024, sae-survey_2025_shu, sae-v_2025} are ideally suited for this purpose, as they are inherently designed to capture interpretable concepts.
We therefore introduce \textbf{\modelname{}}, an SAE-driven interpretable segmentation model that explicitly aligns CoT reasoning with segmentation.
\modelname{} operates end-to-end. Concretely, given an image-instruction pair, the model first reasons to produce a CoT trace and multi-object concentration tokens.
The SAE then maps both the visual tokens and the chain of thought into a shared high-dimensional \emph{sparse feature space}. These features robustly encode semantically rich concepts from both the image and the reasoning trace.
A query codebook selects the salient concepts and aggregates these sparse features into multi-object concept representations.
Next, a slot mapper~\citep{ocl-slot_2020, adaptive-slot_2024} allows the concentration tokens and these concept representations to attend to image features, yielding a multi-slot heatmap that a decoder converts into the final segmentation masks.
This entire mechanism enables the SAE to both explain the reasoning process and guide the segmentation head, offering stronger interpretability than latent query alignment and a more transparent, coherent alignment than textual localization readout.
Concurrently, we optimize the reasoning path by reinforcement learning and jointly supervise the segmentation path, unifying both components under a single training objective.

Across standard benchmark datasets, including gRefCOCO~\citep{ReLA_2023}, RefCOCO~\citep{refcoco_withplus_2014}, RefCOCO+, RefCOCOg~\citep{refcocog_2016}, and ReasonSeg~\citep{lisa_2024}, \modelname{} matches or surpasses best results under comparable evaluation protocols.
Furthermore, our qualitative and quantitative analyses of interpretability reveal a strong correlation between the sparse concept features and the predicted masks.
These extensive experimental results demonstrate that \modelname{} is not only high-performing but also genuinely interpretable, validating that the SAE-derived sparse concepts and their codebook aggregation consistently improve mask quality.

\begin{figure*}[t]
    \captionsetup{aboveskip=4pt, belowskip=-11pt}
    \centering
    \includegraphics[width=\textwidth]{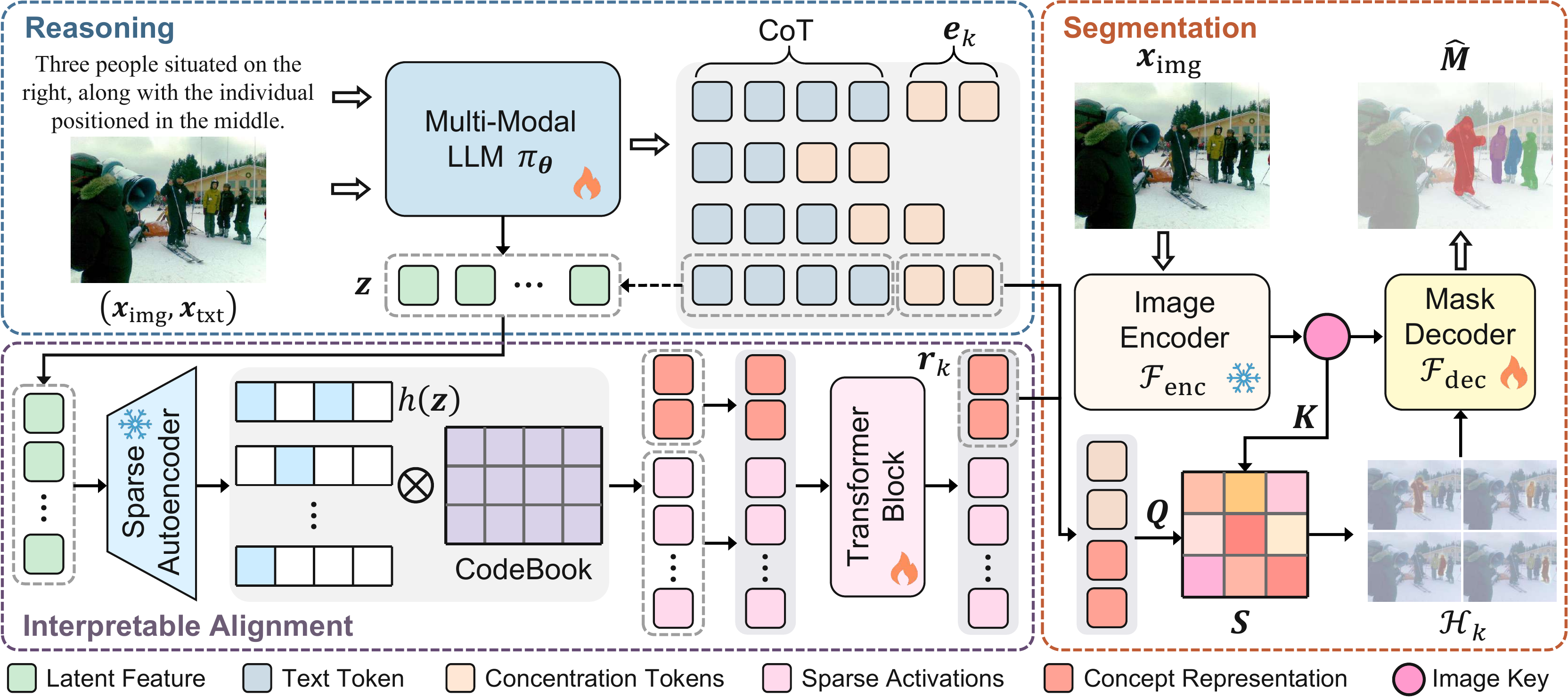}
    \caption{\textbf{\modelname{} pipeline.} Given input, the policy generates CoT and concentration tokens.
    SAE maps the reasoning process into a high-dimensional sparse concept space, and subsequent modules encode it into concept representations.
    Both concentration tokens and concept representations act as queries that attend to image keys, producing a multi-slot heatmap that is decoded into the predicted masks.
    }
    \label{fig:arch}
\end{figure*}

We summarize our contributions as follows.
\setlist{nolistsep}
\begin{itemize}
\item \textbf{Novel Formulation.} We propose \textbf{\modelname{}}, a new end-to-end architecture that, for the first time, integrates a SAE to forge an explicit and interpretable link between CoT reasoning and dense visual segmentation.

\item \textbf{Sparse Concept Interface.} We design an SAE-driven sparse concept interface, which maps both language reasoning traces and visual tokens into a shared, high-dimensional, and informative concept space, creating a transparent pathway from reasoning to mask generation.

\item \textbf{Unified Optimization Strategy.} We successfully unify the reasoning path (trained via reinforcement learning) and the segmentation path (trained via supervised learning) under an end-to-end optimization stage.

\item \textbf{Promising Performance with Interpretability.} 
Our model matches or surpasses top methods on 5 challenging benchmarks. Qualitative and quantitative analysis demonstrates interpretability insights.
\end{itemize}
\section{Related Work}
\label{sec:related_work}
\subsection{Referring and Reasoning Segmentation}
\label{subsec:referring_reasoning}
Referring segmentation localizes target objects based on a short instruction, while reasoning segmentation requires compositional understanding. Early methods~\citep{VLT_2021, CRIS_2022, LAVT_2022, X-Decoder_2023, SEEM_2023, ReLA_2023} typically rely on specific text encoders to parse the text and predict the mask. Recently, LISA~\citep{lisa_2024} bridges the gap between multimodal large language models~(MLLMs) and reasoning segmentation by introducing a special token. Subsequent works, including PerceptionGPT~\citep{perceptiongpt_2024}, PixelLM~\citep{pixellm_2024}, SegLLM~\citep{segllm_2025}, OMG-LLaVA~\citep{OMG-LLaVA_2024}, GroundHog~\citep{GroundHog_2024}, GLaMM~\citep{GLaMM_2024}, UniPixel~\citep{UniPixel}, UniRES~\citep{UFO}, LIRA~\citep{LIRA}, RAS~\citep{RAS_2025}, follow the same paradigm, utilizing LLM latent features and decoding them into segmentation masks. In addition, some works, such as SAM4MLLM~\citep{sam4mllm_2024}, Seg-Zero~\citep{segzero_2025}, Seg-R1~\citep{seg-r1_2025}, and VisionReasoner~\citep{VisionReasoner}, use MLLMs to generate textual coordinates of boxes and points via chain-of-thought, and then feed them to SAM for mask prediction. In a similar vein, Text4Seg~\citep{text4seg_2025} generates textual patch indices and applies CRF~\citep{CRF_2011} or SAM for mask refinement.
However, existing methods either rely on opaque latent query alignments or textual localization that expose rationales but fail to provide explicit, interpretable spatial cues, leaving a gap between reasoning and mask generation. 
In contrast, our \modelname{} bridges reasoning and segmentation through an interpretable pathway that maps reasoning outputs to mask predictions using sparse concept encoding and GRPO~\citep{grpo_2024} training.

\subsection{Sparse Autoencoders}
\label{subsec:sparse_autoencoders}
Sparse Autoencoders (SAEs)~\citep{sae_2024, scaling-sae_2024, sae_base_2025, sae-survey_2025_shu} interpret the internal mechanism of LLMs by disentangling the complex, superimposed features within LLMs into more interpretable components. Specifically, an SAE, trained to reconstruct activations of a target layer under sparsity constraints, extracts a richer set of single-activation features, offering a finer-grained basis for concept interpretation and enabling more detailed analysis of the model’s internal representations and learned knowledge. Prior works~\citep{sae_know_entity, sae_instruction_following, sae_steering_vectors, sam_interpreting} have employed SAEs to analyze the model behavior and control the model output related to unsafe content. Recently, SAE-V~\citep{sae-v_2025} extends the SAE paradigm to multi-modal LLMs, yielding more interpretable and disentangled latent representations. We build on this insight by aligning interpretable representations with downstream segmentation tasks, enabling a transparent connection from reasoning outputs to mask generation.

\section{Method}
\label{sec:method}

\subsection{Architecture}
\label{subsec:architecture}

\minihead{Overall.}
As shown in \cref{fig:arch}, \modelname{} aligns reasoning and segmentation through a sparse-concept pathway.
Given an image-instruction pair~$\left( \ximg, \, \xtxt \right)$, the MLLM policy~$\policy \! \left( \cdot \right)$ generates a token sequence~$\yoneT$ that contains the chain-of-thought~(CoT) and $K_s$ concentration tokens.
We read its hidden states to obtain the embeddings of concentration tokens.
The sparse autoencoder~(SAE)~\citep{sae-v_2025} encodes the hidden states from $( \ximg, \, \xtxt, \, \yoneT )$ at a chosen layer, into a high-dimensional sparse space.
Then the query codebook $\bm{C} \in \mathbb{R} ^ {d_\text{sae} \times d_c}$ (with $d_\text{sae}$ the SAE dimension and $d_c$ the concept representation dimension) selects these sparse features, and Transformer~\citep{attention_2017} encoder blocks aggregate them into $K_s$ concept representations.
In parallel, The vision encoder~$\Fenc \! \left( \cdot \right)$ extracts image features to serve as image keys.
The concentration token embeddings and concept representations are fused into $K_s$ queries, which attend over the image keys and yield an observable $K_s \text{-slot}$ heatmap with confidences~$\heatconf$ via the slot mapper~$\Fslot \! \left( \cdot \right)$.
Finally, the mask decoder~$\Fdec \! \left( \cdot \right)$ converts the heatmap into the predicted masks~$\hat{\bm{M}}$. \par

\minihead{Reasoning Module.}
We use LLaVA-1.5~\citep{llava1.5_2024} and Qwen2.5-VL~\citep{qwen2.5_vl} as our MLLM backbone.
Following DeepSeek-R1~\citep{deepseek-r1_2025}, we adopt multi-modal chain-of-thought~(MCoT)~\cite{MCoT_Survey_2025} to leverage the reasoning capabilities of MLLMs on compositional instructions.
Specifically, we use an instruction prompt to elicit both the CoT and $K_s$ concentration tokens.
Given $\left( \ximg, \, \xtxt \right)$, the model is asked to (i) reason in a \texttt{<think>...</think>} block and then (ii) output $K_s$ concentration tokens, where $K_s$ is a hyper-parameter.
Under this setup, the policy $\policy$ generates the token sequence $\yoneT$ via next token prediction.
Formally, the process is given in
{%
  \SetEqBlank
  \begin{equation}
    \begin{gathered}
      y_{t} \sim \policy \! \left( \cdot \mid \bm{y}_{\scriptscriptstyle 0:t-1},\, \ximg,\, \xtxt \right), \;
      t=1,\ldots,T,\\
    \end{gathered}
  \end{equation}
}\noindent\ignorespaces
where $\yoneT$ includes both the CoT and the concentration tokens.
We obtain the $K_s$ concentration tokens' embeddings $\ConcenEmbed$ by locating their positions in the sequence and reading the corresponding hidden states from the MLLM.

\minihead{Interpretable Alignment (Overview).}
We pass the image–text prefill together with CoT through an SAE, which encodes the problem statement and reasoning process into sparse concept features.
A learned query codebook then selects salient concepts, and Transformer encoder blocks aggregate them into $K_s$ concept representations $\ConceptRepre$.
Next, we fuse the concentration token embeddings with the concept representations using a lightweight MLP to form slot queries $\bm{Q}\!\in\!\mathbb{R}^{K_s \times d_q}$.
In parallel, the vision backbone encodes $\ximg$ into image keys $\bm{K}\!\in\!\mathbb{R}^{h\times w\times d_k}$. In practice we adopt ViT-H (the SAM~\citep{sam_2023} image encoder) as the vision backbone.
 Finally, a slot mapper~$\Fslot \left( \cdot \right)$ operates on $\bm{Q}$ and $\bm{K}$, computing multi-head attention scores and aggregating features across heads to produce a multi-slot heatmap $\heatmap$ and per-slot confidences $\confidence$.
Formally, the process is given by
 {%
  \SetEqBlank
  \label{eq:ComputeQK}
  \begin{equation}
    \begin{gathered}
      \bm{K} = \mathcal{F}_{\text{enc}} \! \left( \ximg \right), \\
      \bm{Q}_k = \mathrm{MLP}\, \big(\mathrm{Concat} (\bm{e}_k, \, \bm{r}_k) \big), \;
      k=1,\ldots,K_s,\\
      \heatconf = \Fslot \left(  \bm{Q}, \, \bm{K} \right).
    \end{gathered}
  \end{equation}
}\noindent\ignorespaces

\minihead{Segmentation Decoder.}\
The decoder has two lightweight components.
First, three stacked 2D convolutional blocks resample the heatmap to the decoder resolution, yielding a feature map.
Second, following the SAM~\citep{sam_2023} decoder design, we employ a Two-Way Transformer to perform bidirectional cross-attention between image keys and the feature map.
Formally, the process is
{%
\SetEqBlank
  \begin{equation}
    \hat{\bm{M}} = \mathcal{F}_{\text{dec}} \! \left( \bm{K}, \, \heatconf \right).
  \end{equation}
}\noindent\ignorespaces

\subsection{Interpretable Alignment}
\label{subsec:inter_align}

\minihead{Sparse Autoencoder.}A sparse autoencoder~(SAE)~\citep{sae-v_2025} factorizes token representations into a small set of sparse interpretable features.
For a sequence processed by an LLM, we take tokens' hidden states $\bm{z} \in \mathbb{R} ^ {T \times d_{ \pi}}$ at a chosen layer, where $T$ is the sequence length and $d_{ \pi}$ is the hidden dimension of LLM policy (notation follows \cref{subsec:architecture}).
The SAE encoder~$\Esae \left( \cdot \right)$ applies a linear map followed by a sparsifying activation to produce high-dimensional sparse activations $h (\bm{z}) \in \mathbb{R}^{T \times d_{\text{sae}}}$ with $d_{\text{sae}}\gg d_{ \pi}$ (e.g., $d_{\text{sae}}=65536$ and $d_{ \pi}=4096$).
Each coordinate acts as a dictionary atom that activates only when its concept is present.
The decoder~$\Dsae \left( \cdot \right)$ is a linear map that projects $h (\bm{z})$ back to $\hat {\bm {z}}$ in the original space, reconstructing $\bm{z}$.
This overcomplete sparse coding promotes feature disentanglement and makes features easy to select or aggregate downstream, and it applies uniformly to text and vision tokens.

\begin{figure}[ht]
\captionsetup{aboveskip=6pt, belowskip=-8pt}
    \centering
    \includegraphics[width=\columnwidth]{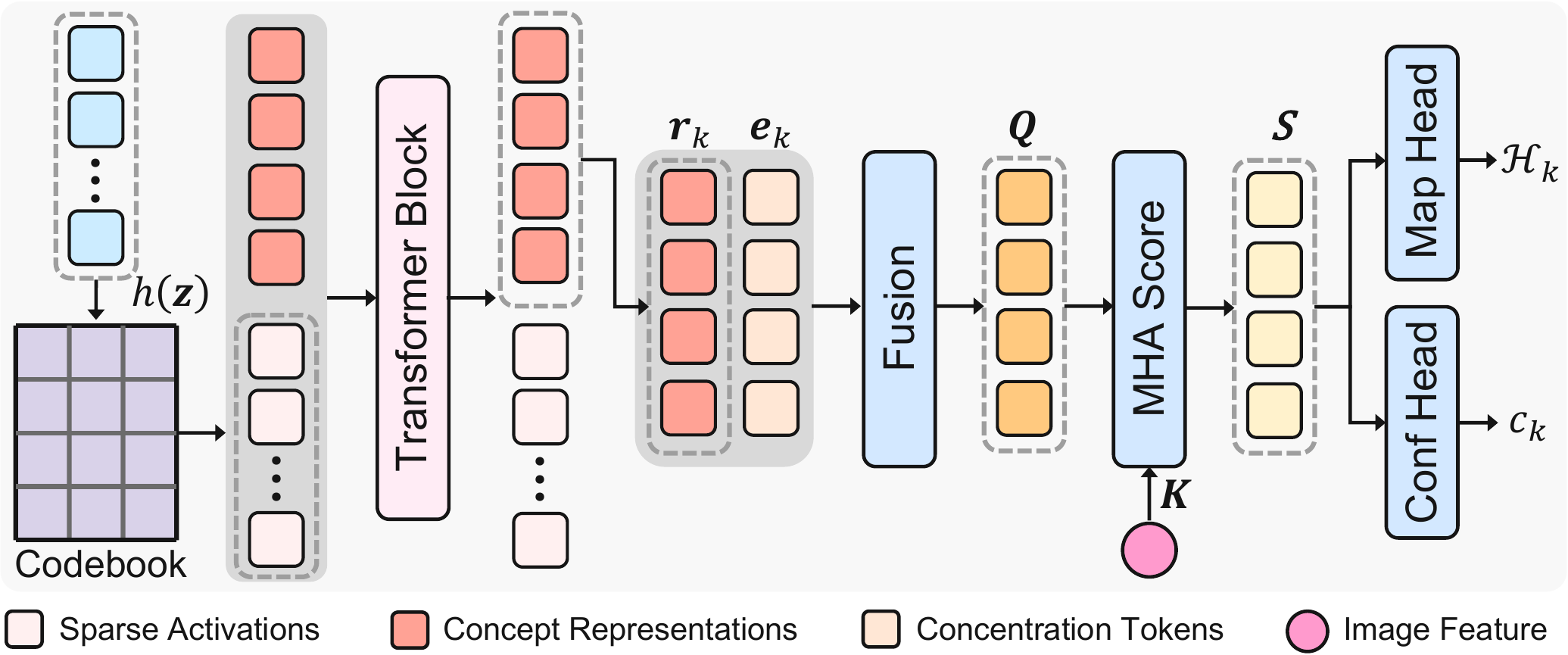}
    \caption{\textbf{Detailed architecture} from the query codebook to the slot mapper. Inputs: sparse features from the SAE. Outputs: a multi-slot heatmap with per-slot confidences.}
    \label{fig:seg_architecture}
\end{figure}

\minihead{Query Codebook \& Slot Mapper.}
As shown in~\cref{fig:seg_architecture}, we apply a query codebook with Transformer encoder blocks to select and aggregate $h (\bm{z})$, and use a slot mapper to yield multi-slot heatmap with confidences.
Specifically, we first filter nonzero activations in the sparse concept space to obtain index–concept pairs $\{ \left(j, h_j(\bm{z}) \right) \} _{j \in \SuppZ }$, where the support set is defined in
{%
\SetEqBlank
  \begin{equation}
   \SuppZ \;=\; \left\{ j \in \{1,\dots,d_{\text{sae}}\} : h_j(\bm{z}) \neq 0 \right\}.
  \end{equation}
}\noindent\ignorespaces
Moreover, $\lvert \SuppZ \lvert \ll d_\text{sae}$ (see analysis in \cref{subsec:interp_analysis}).
Second, we use the codebook $\bm{C} \in \mathbb{R}^{d_\text{sae} \times d_c }$ to decode $\{ \left(j, h_j(\bm{z}) \right) \} _{j \in \SuppZ }$ into the dense space.
Then we initialize $K_s$ concept representations and, together with $  \{ \bm{C} ( h_j(\bm{z}) ) \} _{j\in\SuppZ} $, pass them through Transformer encoder blocks with self-attention to obtain $\ConceptRepre$.
Finally, using \cref{eq:ComputeQK} to form the queries~$Q$ and keys~$K$, we compute the multi-slot heatmap and the corresponding confidences as in
{%
  \SetEqBlank
  \begin{equation}
  \begin{gathered}
  \bm{S} = \big[ (\bm{Q}\bm{W}^{Q}_{i})(\bm{K}\bm{W}^{K}_{i})^{\top}/\sqrt{d_h} \, \big]_{i=1}^{N_h}, \\
  \heatmap = \mathcal{F}_{\text{map}} (\bm{S}), \; \;
  \confidence = \mathcal{F}_{\text{conf}} (\bm{S}),
  \end{gathered}
  \end{equation}
}\noindent\ignorespaces
where $\bm{Q}\!\in\!\mathbb{R}^{K_s \times d_q}$, $\bm{K}\!\in\!\mathbb{R}^{h\times w\times d_k}$,  $\bm{W}^{Q}_{i}\!\in\!\mathbb{R}^{d_q\times d_h}$, $\bm{W}^{K}_{i}\!\in\!\mathbb{R}^{d_k\times d_h}$, $d_h$ is the head dimension, $N_h$ is the number of heads, the map head $\mathcal{F}_{\text{map}}:\mathbb{R}^{N_h \times K_s \times h\times w}\!\to\!\mathbb{R}^{K_s \times h\times w}$, 
and the confidence head $\mathcal{F}_{\text{conf}}:\mathbb{R}^{N_h \times K_s \times h\times w}\!\to\!\mathbb{R}^{K_s}$.

\minihead{Interpretable Pathway.}
The alignment is interpretable end-to-end.
The SAE surfaces active dictionary atoms as explicit index–activation pairs ${(j,h_j(\bm z))}_{j\in\SuppZ}$, yielding concept-level attribution in a high-dimensional sparse space.
A learned query codebook and Transformer encoder blocks then select and consolidate these sparse concepts into $K_s$ concept representations $\ConceptRepre$ while preserving provenance (which indices from $\SuppZ$ contributed and with what weights).
Finally, the slot mapper $\Fslot$ maps the resulting queries and vision keys to an observable multi-slot heatmap $\heatmap$ and per-slot confidences $\confidence$, making each slot’s spatial footprint and reliability directly inspectable.
Detailed analyses are provided in \cref{subsec:interp_analysis}.

\subsection{Learning Objectives}
\label{subsec:learning_obj}

\minihead{Overall.}
We train the whole system with the objective that couples reinforcement learning~\citep{grpo_2024} on the language path with segmentation supervision on the vision path.
For each $\left( \ximg, \, \xtxt \right)$, the policy~$\policy$ rolls out a group of responses~$\bigl\{ \yi \bigr\} _{i=1} ^ {G}$ with the group size~$G$, and we compute the GRPO~\citep{grpo_2024} loss~$\Lgrpo$ from the advantages.
In parallel, the multi-slot heatmap and confidences~$\heatconf$ are matched with the ground truth masks~$\Mgt$.
 $\Mgt$ supervises $\heatmap$ and $\hat{\bm{M}}$, while the binary targets $\confidenceGT$ derived from the matching supervise $\confidence$.
Formally, with coefficients $\lambda_{\textsc{s}}$ and $\lambda_{\textsc{c}}$, the overall objective is given in
{%
\SetEqBlank  
\begin{equation}
\begin{aligned}
  \mathcal{L} \, =  & \, \Lgrpo \Bigl( \bigl\{ \yi \bigl\} _ {i=1}^{G} \Bigr) + \lambda_{\textsc{s}} \Lseg \Bigl( \heatmap, \, \hat{\bm{M}}, \, \Mgt \Bigr) \\
  & + \lambda_{\textsc{c}} \Lconf \Bigl( \confidence, \, \confidenceGT \Bigr),
\end{aligned}
\end{equation}
}\noindent\ignorespaces
where $(\cdot)$ denotes ordered sequences and $\{\cdot\}$ denotes unordered sets. Each loss term is specified in the following paragraphs.

\minihead{SAE Learning Objective.}
We pretrain the SAE to reconstruct token representations while enforcing sparsity in the activation space.
Given hidden states $\bm{z}$ and the encoded sparse activations $h(\bm{z}) = \Esae \left( \bm{z} \right)$,  the objective balances reconstruction fidelity and sparsity:
{%
\SetEqBlank
\begin{equation}
\Lsae (\bm{z}) = \Vert \bm{z} - \hat{\bm{z}}\Vert _2 ^2
	\, + \, \alpha \Vert h(\bm{z}) \Vert_{1}.
\label{eq:sae_loss}
\end{equation}
}\noindent\ignorespaces
Here $\hat{\bm{z}} = \Dsae \big(h(\bm{z})\big)$ reconstructs $\bm{z}$ to preserve information.
The $\ell_1$ penalty drives most coordinates of $h(\bm{z})$ to zero, yielding a compact concept basis, and $\alpha$ is a control factor.

\minihead{Reinforcement Learning Objective.}
Following \citet{grpo_2024}, we optimize the policy $\policy$ with the GRPO objective.
We specify the GRPO loss $\Lgrpo$ in the Appendix.
For each sampled response in a group, we compute a segmentation reward and a format reward.
For segmentation reward, we perform bipartite matching between predicted masks~$\hat{\bm{M}}$ and ground truth masks~$\Mgt$, and assign each matched pair a score that combines predicted confidence and mask IoU.
We score CoT formatting via a set of regular-expression checks as the format reward.
Both rewards are normalized to $[0, 1]$ with fixed coefficients.
Further implementation details are provided in Appendix.

\minihead{Supervised Segmentation Objectives.}
The segmentation objectives comprise a segmentation loss and a confidence loss.
Losses are computed only on matched pairs between predictions and ground truths, via bipartite matching. 
First, a binary cross entropy~(BCE) loss on the multi-slot heatmap $\heatmap$ encourages concentrated spatial evidence. Second, a dice loss~\cite{dice_loss} on the predicted masks $\hat{\bm{M}}$ directly supervises mask quality with the coefficient $\lambda_\textsc{d}$.
Using the same matching, slot confidences $\confidence$ are trained with BCE against the binary targets $\confidenceGT \in \{ 0,1 \}^{K_s}$ (1 if the slot is matched to any ground truth instance, 0 otherwise).
Formally, the segmentation objectives are
{%
\SetEqBlank
\begin{equation*}
\begin{gathered}
  \Lseg \,=\, \mathcal{L}_{\text{\scriptsize \textsc{bce}}}
  \bigl( \heatmap ,\, \Mgt \bigr)
  + \lambda_\textsc{d} \mathcal{L}_{\text{\scriptsize \textsc{dice}}}
  \bigl( \hat{\bm{M}} ,\, \Mgt \bigr), \\
  \Lconf \,=\, \frac{1}{K_s} \sum_{k=1} ^ {K_s}\mathcal{L}_{\text{\scriptsize \textsc{bce}}}
  \bigl( c_k, \, y_k \bigr).
\end{gathered}
\end{equation*}
}\noindent\ignorespaces

\section{Experiments}
\label{sec:experiments}

\minihead{Research Questions.}
We aim to answer the following research questions in this section:

\begin{description}[labelindent=.5em]
  \item[\textbf{RQ1:}] Does the model achieve higher accuracy in reasoning segmentation and state-of-the-art results on standard benchmarks relative to prior methods?
  \item[\textbf{RQ2:}] How does the model’s interpretability manifest, and does qualitative and quantitative evidence support the claim that the SAE explains the model’s reasoning process?
  \item[\textbf{RQ3:}] How do the reinforcement learning and segmentation supervision contribute to performance, how does the choice of vision backbone affect results, and how sensitive is the model to GRPO hyperparameters?
\end{description}

\subsection{Experimental Setup}
\label{subsec:setup}
\minihead{Backbones, Datasets and Metrics.}
Our experiments use three MLLM backbones: Qwen2.5-VL-7B~\citep{qwen2.5_vl}, LLaVA-1.5-7B~\cite{llava1.5_2024}, and LLaVA-1.5-13B.
Using 200K samples from OBELICS~\citep{obelics_2023}, we pretrain SAEs separately for each backbone.
We train \modelname{} with each backbone on the training sets of RefCOCO~\citep{refcoco_withplus_2014}, RefCOCO+, RefCOCOg~\citep{refcocog_2016}, and gRefCOCO (multi-object)~\citep{ReLA_2023}.
We evaluate on the official validation and test splits of RefCOCO(+/g) and gRefCOCO. 
We further assess zero-shot reasoning segmentation by evaluating on ReasonSeg (validation and test) without using its images for training.
Following common practice, we report cIoU~(the cumulative intersection over the cumulative union) on RefCOCO(+/g), and both cIoU and gIoU~(mean of per-image IoU) on gRefCOCO and ReasonSeg.

\minihead{Baselines.}
We compare our method with 27 prior works, grouped into 3 categories according to their architecture.
Methods without LLMs include methods that do not rely on LLMs to encode textual instructions for mask generation, such as VLT~\citep{VLT_2021}, CRIS~\citep{CRIS_2022}, LAVT~\citep{LAVT_2022}, ReLA~\citep{ReLA_2023}, X-Decoder~\citep{X-Decoder_2023}, and SEEM~\citep{SEEM_2023}. 
Latent Query Alignment Methods, including LISA~\citep{lisa_2024}, PerceptionGPT~\citep{perceptiongpt_2024}, PixelLM~\citep{pixellm_2024}, LaSagnA~\citep{lasagna_2024}, OMG-LLaVA~\citep{OMG-LLaVA_2024}, GroundHog~\citep{GroundHog_2024}, GLaMM~\citep{GLaMM_2024}, SegLLM~\citep{segllm_2025}, UniPixel~\citep{UniPixel}, UniRES~\citep{UniRES}, UFO~\citep{UFO}, LIRA~\citep{LIRA}, HiMTok~\citep{himtok_2025}, RAS~\citep{RAS_2025}, and X-SAM~\citep{XSAM}, typically take hidden features or queries from a large language model and decode them into segmentation masks. 
Textual Localization Readout Methods, which includes SAM4MLLM~\citep{sam4mllm_2024}, Seg-R1-7B~\citep{seg-r1_2025}, Seg-Zero-7B~\citep{segzero_2025}, VisionReasoner-7B~\citep{VisionReasoner}, SAM R1-7B~\citep{SAM-R1}, and Text4Seg~\citep{text4seg_2025}, use an MLLM to emit discrete location tokens (e.g., box coordinates or patch indices) which are then converted into masks.

\begin{table}[t]
\captionsetup{skip=3pt}
\centering
\scriptsize
\setlength{\tabcolsep}{3pt} % 原来是14pt，单栏一般需要更紧
\caption{Comparison of methods on RefCOCO, RefCOCO+, and RefCOCOg datasets. 
Backbones: 
$^{1}$ LLaVA-1.5, 
$^{2}$ Qwen2.5-VL, 
$^{3}$ InternVL-2.5, 
$^{4}$ InternVL-2, 
$^{5}$ Phi-3, 
$^{6}$ LLaVA-1.6.}
\label{tab:main_rst_refcoco}
\begin{tabular}{l ccc ccc cc}
\toprule

\multirow{2}{*}{Method} & \multicolumn{3}{c}{RefCOCO} & \multicolumn{3}{c}{RefCOCO+} & \multicolumn{2}{c}{RefCOCOg}\\
\cmidrule(r){2-4} \cmidrule(lr){5-7} \cmidrule(l){8-9}
& val & testA & testB & val & testA & testB & val & test\\
\midrule
\multicolumn{9}{l}{\textit{Method without LLMs}}\\
VLT~\citep{VLT_2021} & 67.5 & 70.5 & 65.2 & 56.3 & 61.0 & 50.1 & 55.0 & 57.7 \\
CRIS~\citep{CRIS_2022} & 70.5 & 73.2 & 66.1 & 62.3 & 68.1 & 53.7 & 59.9 & 60.4 \\
LAVT~\citep{LAVT_2022} & 72.7 & 75.8 & 68.8 & 62.1 & 68.4 & 55.1 & 61.2 & 62.1 \\
ReLA~\citep{ReLA_2023} & 73.8 & 76.5 & 70.2 & 66.0 & 71.0 & 57.7 & 65.0 & 66.0 \\
X-Decoder~\citep{X-Decoder_2023} & -- & -- & -- & -- & -- & -- & 64.6 & -- \\
SEEM~\citep{SEEM_2023} & -- & -- & -- & -- & -- & -- & 65.7 & -- \\
\midrule
\multicolumn{9}{l}{\textit{Latent Query Alignment Method}}\\
LISA-7B$^{1}$~\citep{lisa_2024} & 74.9 & 79.1 & 72.3 & 65.1 & 70.8 & 58.1 & 67.9 & 70.6\\
LISA-13B$^{1}$~\citep{lisa_2024} & 76.0 & 78.8 & 72.9 & 65.0 & 70.2 & 58.1 & 69.5 & 70.5\\
PerceptionGPT-7B$^{1}$~\citep{perceptiongpt_2024}& 75.1 & 78.6 & 71.7 & 68.5 & 73.9 & 61.3 & 70.3 & 71.7\\
PerceptionGPT-13B$^{1}$~\citep{perceptiongpt_2024} & 75.3 & 79.1 & 72.1 & 68.9 & 74.0 & 61.9 & 70.7 & 71.9\\
PixelLM-7B$^{1}$~\citep{pixellm_2024} & 73.0 & 76.5 & 68.2 & 66.3 & 71.7 & 58.3 & 69.3 & 70.5 \\
LaSagnA-7B$^{1}$~\citep{lasagna_2024} & 76.8 & 78.7 & 73.8 & 66.4 & 70.6 & 60.1 & 70.6 & 71.9 \\
OMG-LLaVA-7B$^{1}$~\citep{OMG-LLaVA_2024} & 78.0 & 80.3 & 74.1 & 69.1 & 73.1 & 63.0 & 72.9 & 72.9\\
GroundHog-7B$^{1}$~\citep{GroundHog_2024} & 78.5 & 79.9 & 75.7 & 70.5 & 75.0 & 64.9 & 74.1 & 74.6\\
GLaMM-7B$^{1}$~\citep{GLaMM_2024} & 79.5 & 83.2 & 76.9 & 72.6 & 78.7 & 64.6 & 74.2 & 74.9\\
SegLLM-7B$^{1}$~\citep{segllm_2025} & 80.2 & 81.5 & 75.4 & 70.3 & 73.0 & 62.5 & 72.6 & 73.6 \\
UniPixel-7B$^{2}$~\citep{UniPixel} & 80.8 & 83.0 & 77.4 & 75.3 & 80.1 & 70.0 & 76.4 & 77.1\\
UniRES-7B$^{1}$~\citep{UniRES} & 80.2 & 81.8 & 75.8 & 71.6 & 76.0 & 64.4 & 73.8 & 74.1\\
UFO-8B$^{3}$~\citep{UFO} & 81.0 & 82.6 & 78.6 & 77.1 & 80.4 & 72.6 & 76.7 & 77.3\\
LIRA-8B$^{4}$~\citep{LIRA} & 81.8 & 83.4 & 78.1 & 76.3 & 81.1 & 70.5 & 78.4 & 78.2\\
HiMTok-8B$^{3}$~\citep{himtok_2025} & \textit{85.9} & 86.3 & \underline{83.9} & \textbf{80.5} & 83.7 & \textit{76.4} & 80.1 & 80.9 \\
RAS-13B$^{1}$~\citep{RAS_2025} & 81.0 & 83.5 & 79.0 & 75.1 & 80.0 & 70.3 & 76.0 & 77.5\\
X-SAM-3.8B$^{5}$~\citep{XSAM} & 85.1 & \textit{87.1} & 83.4 & 78.0 & 81.0 & 74.4 & \underline{83.8} & \textbf{83.9}\\
\midrule
\multicolumn{9}{l}{\textit{Textual Localization Readout Method}}\\
SAM4MLLM-7B$^{6}$~\citep{sam4mllm_2024} & 79.6 & 82.8 & 76.1 & 73.5 & 77.8 & 65.8 & 74.5 & 75.6\\
Seg-R1-7B$^{2}$~\citep{seg-r1_2025} & 74.3 & 78.7 & 67.6 & 62.6 & 70.9 & 57.9 & 71.0 & 71.4\\
Seg-Zero-7B$^{2}$~\citep{segzero_2025} & -- & 80.3 & -- & -- & 76.2 & -- & -- & 72.6\\
VisionReasoner-7B$^{2}$~\citep{VisionReasoner} & -- & 78.9 & -- & -- & 74.9 & -- & -- & 71.3\\
SAM-R1-7B$^{2}$~\citep{SAM-R1} & -- & 79.2 & -- & -- & 74.7 & -- & -- & 73.1\\
Text4Seg-7B$^{1}$~\citep{text4seg_2025} & 79.3 & 81.9 & 76.2 & 72.1 & 77.6 & 66.1 & 72.1 & 73.9\\
Text4Seg-13B$^{1}$~\citep{text4seg_2025} & 80.2 & 82.7 & 77.3 & 73.7 & 78.6 & 67.6 & 74.0 & 75.1 \\
\midrule
\multicolumn{9}{l}{\textit{Interpretable Alignment Method}}\\
\rowcolor{gray!20}
\modelname{}-7B$^{1}$ & 80.0 & 82.9 & 79.2 & 77.2 & \textit{83.9} & 71.8 & 79.4 & 77.6 \\
\rowcolor{gray!20}
\modelname{}-7B$^{2}$ & \underline{85.3} & \textbf{87.5} & \textit{83.6} & \underline{78.3} & \textbf{84.7} & \underline{76.6} & \textit{82.8} & \textit{81.0} \\
\rowcolor{gray!20}
\modelname{}-13B$^{1}$ & \textbf{86.3} & \underline{87.3} &  \textbf{84.1} & \textbf{80.5} & \underline{84.6} & \textbf{76.9} & \textbf{84.0} & \underline{81.3} \\

\bottomrule
\end{tabular}
\end{table}

\minihead{Implementation Details.}
For each experiment, we train on 8 NVIDIA A100 (80 GB) GPUs.
For GRPO, we ablate the number of samples at $\{\text{2}, \text{4},  \text{6}, \text{8}\}$.
We weight the two reward components as 0.7 for segmentation and 0.3 for CoT formatting.
For single-object datasets, we use the soft IoU as the segmentation score.
For multi-object datasets, we add a binary confidence score (1 if matched, 0 otherwise).
The format score is computed under specific regular expression rules for five conditions~(see Appendix).
For the slot mapper, we set the maximum number of slots to 6.
Loss coefficients $\lambda_{\textsc{s}}$, $\lambda_{\textsc{c}}$ and $\lambda_{\textsc{d}}$ are set to 1.0, 0.2 and 0.6, respectively.
The base learning rate for the MLLM backbone is set to 2e-6, and we apply multipliers of $25\times$ for the query codebook, $80 \times$ for the slot mapper, and $10\times$ for mask decoder.
We use the AdamW~\citep{adamw} optimizer with weight decay 0.01 and adopt OneCycleLR~\citep{one_cycle_Lr} as the learning rate scheduler. 
Full configurations are provided in Appendix.
Selections of significant hyper-parameters are revealed in \Cref{subsec:ablation}.

\begin{table}[t]
\captionsetup{skip=3pt}
\centering
\scriptsize
\setlength{\tabcolsep}{4pt} 
\caption{Comparison of methods on gRefCOCO. * indicates zero-shot performance.
Backbones: 
$^{1}$ LLaVA-1.5, 
$^{2}$ Qwen2.5-VL, 
$^{3}$ InternVL-2.5, 
$^{4}$ InternVL-2, 
$^{5}$ Phi-3, 
$^{6}$ LLaVA-1.6.}
\label{tab:main_rst_gref}
\begin{tabular}{l c c c c c c c}
\toprule
\multirow{2}{*}{Method} & \multirow{2}{*}{Year} & \multicolumn{2}{c}{val} & \multicolumn{2}{c}{testA} & \multicolumn{2}{c}{testB}\\
\cmidrule(r){3-4} \cmidrule(lr){5-6} \cmidrule(l){7-8}
& & gIoU & cIoU & gIoU & cIoU & gIoU & cIoU\\
\midrule
\multicolumn{8}{l}{\textit{Method without LLMs}}\\
MattNet~\citep{MattNet} &\citeyear{MattNet} & 48.2 & 47.5 & 59.3 & 58.6 & 46.1 & 45.3 \\
LTS~\citep{LTS} &\citeyear{LTS} & 52.7 & 52.3 & 62.6 & 61.8 & 50.4 & 49.9 \\
VLT~\citep{VLT_2021} &\citeyear{VLT_2021} & 52.0 & 52.5 & 63.2 & 62.1 & 50.8 & 50.5 \\
CRIS~\citep{CRIS_2022} &\citeyear{CRIS_2022} & 56.2 & 55.3 & 63.4 & 63.8 & 51.7 & 51.0 \\
LAVT~\citep{LAVT_2022} &\citeyear{LAVT_2022} & 58.4 & 57.6 & 65.9 & 65.3 & 55.8 & 55.0 \\
ReLA~\citep{ReLA_2023} &\citeyear{ReLA_2023} & 63.6 & 62.4 & 70.0 & 69.3 & 61.0 & 59.9 \\
\midrule
\multicolumn{8}{l}{\textit{Latent Query Alignment Method}}\\
LISA-7B$^{1}$~\citep{lisa_2024} &\citeyear{lisa_2024} & 61.6 & 61.7 & 66.2 & 68.5 & 58.8 & 60.6\\
LISA-13B$^{1}$~\citep{lisa_2024} &\citeyear{lisa_2024} & 63.4 & 62.9 & 68.1 & 69.6 & 61.8 & 62.2\\
GSVA-7B$^{1}$~\citep{GSVA} &\citeyear{GSVA} & 66.5 & 63.3 & 71.1 & 69.9 & 62.2 & 60.5\\
GSVA-13B$^{1}$~\citep{GSVA} &\citeyear{GSVA} & 68.0 & 64.0 & 71.8 & 70.5 & 63.8 & 61.3\\
GroundHog-7B$^{1}$~\citep{GroundHog_2024} &\citeyear{GroundHog_2024} & 66.7 & -- & -- & -- & -- & --\\
LaSagnA-7B*$^{1}$~\citep{lasagna_2024} &\citeyear{lasagna_2024} & 32.4 & 38.1 & 47.3 & 50.4 & 38.9 & 42.1 \\
OMG-LLaVA-7B*$^{1}$~\citep{OMG-LLaVA_2024} &\citeyear{OMG-LLaVA_2024} & 36.1 & 39.3 & 50.1 & 52.4 & 42.2 & 43.7\\
LIRA-8B*$^{4}$~\citep{LIRA} &\citeyear{LIRA} & 36.7 & 40.9 & 50.4 & 52.4 & 42.4 & 44.9 \\
HiMTok-8B$^{3}$~\citep{himtok_2025} &\citeyear{himtok_2025} & 72.1 & 70.4 & 73.5 & 74.9 & \underline{71.7} & \textit{72.0} \\
UniRES-7B$^{1}$~\citep{UniRES} &\citeyear{UniRES} & 74.4 & 69.9 & 76.0 & 74.5 & 69.8 & 66.6 \\
RAS-13B$^{1}$~\citep{RAS_2025} &\citeyear{RAS_2025} & 74.6 & 70.5 & \underline{77.5} & \underline{77.0} & 69.4 & 67.9\\
\midrule

\multicolumn{8}{l}{\textit{Textual Localization Readout Method}}\\
SAM4MLLM-7B$^{6}$~\citep{sam4mllm_2024} &\citeyear{sam4mllm_2024} & 71.9 & 67.8 & 74.2 & 72.2 & 65.3 & 63.4\\
Text4Seg-7B$^{1}$~\citep{text4seg_2025} &\citeyear{text4seg_2025} & 73.6 & 67.9 & 74.1 & 72.8 & 66.1 & 64.8\\
Text4Seg-13B$^{1}$~\citep{text4seg_2025} &\citeyear{text4seg_2025} & 74.8 & 69.8 & 75.1 & 74.3 & 68.0 & 67.1\\
\midrule

\multicolumn{8}{l}{\textit{Interpretable Alignment Method}}\\

\rowcolor{gray!20}
\modelname{}-7B$^{1}$ & & \textit{75.0} & \textit{70.6} & 76.1 & 72.6 & 68.2 & 66.7 \\
\rowcolor{gray!20}
\modelname{}-7B$^{2}$ & & \underline{76.1} & \textbf{72.2} & \textit{76.8} & \textit{76.5} & 70.0 & \textbf{72.8} \\
\rowcolor{gray!20}
\modelname{}-13B$^{1}$ & & \textbf{76.8} & \textbf{72.2} & \textbf{77.7} & \textbf{77.3} & \textbf{71.9} & \underline{72.4} \\
\bottomrule
\end{tabular}
\end{table}

\subsection{Overall Performance (RQ1)}
\label{subsec:overll_performance}

We evaluate the effectiveness of \modelname{} by conducting experiments on three standard benchmarks: the RefCOCO series, gRefCOCO, and ReasonSeg.
% Refcoco series Result
\minihead{Results on RefCOCO(+/g).} Following the evaluation protocols of prior reasoning segmentation methods~\cite{lisa_2024, RAS_2025}, we conduct experiments on the RefCOCO series. As shown in \Cref{tab:main_rst_refcoco}, our \modelname{} achieves the best performance on almost all splits. Moreover, under the same base model and model size conditions, our model achieves significant performance improvements, demonstrating the effectiveness. In addition, compared to Seg-Zero, SAM-R1, and VisionReasoner trained via GRPO, \modelname{} performs better. This advantage stems from our designed sparse feature interface that applies sparse concept features to align thought reasoning with segmentation.

\begin{table}[t]
\captionsetup{skip=3pt}
\centering
\scriptsize
\setlength{\tabcolsep}{7.5pt} 
\caption{Zero-shot comparison of methods on ReasonSeg dataset.
Backbones: 
$^{1}$ LLaVA-1.5, 
$^{2}$ Qwen2.5-VL, 
$^{3}$ InternVL-2.5, 
$^{4}$ InternVL-2, 
$^{5}$ Phi-3, 
$^{6}$ LLaVA-1.6.}
\label{tab:main_rst_reasonseg}
\begin{tabular}{l c c c c c}
\toprule
\multirow{2}{*}{Method} & \multirow{2}{*}{Year} & \multicolumn{2}{c}{val} & \multicolumn{2}{c}{test} \\
\cmidrule(r){3-4} \cmidrule(l){5-6}
& & gIoU & cIoU & gIoU & cIoU \\
\midrule

\multicolumn{6}{l}{\textit{Method without LLMs}}\\
ReLA~\citep{ReLA_2023} &\citeyear{ReLA_2023} & 22.4 & 19.9 & 21.3 & 22.0 \\
X-Decoder~\citep{X-Decoder_2023} &\citeyear{X-Decoder_2023} & 22.6 & 17.9 & 21.7 & 16.3 \\
SEEM~\citep{SEEM_2023} &\citeyear{SEEM_2023} & 25.5 & 21.2 & 24.3 & 18.7 \\
Grounded-SAM~\citep{groundedsam_2024} &\citeyear{groundedsam_2024} & 26.0 & 14.5 & 21.3 & 16.4 \\
\midrule

\multicolumn{6}{l}{\textit{Latent Query Alignment Method}}\\
LISA-7B$^{1}$~\citep{lisa_2024} &\citeyear{lisa_2024} & 53.6 & 52.3 & 48.7 & 48.8 \\
LISA-13B$^{1}$~\citep{lisa_2024} &\citeyear{lisa_2024} & 57.7 & 60.3 & 53.8 & 50.8 \\
LaSagnA-7B$^{1}$~\citep{lasagna_2024} &\citeyear{lasagna_2024} & 48.8 & 47.2 & -- & -- \\
GroundHog-7B$^{1}$~\citep{GroundHog_2024} &\citeyear{GroundHog_2024} & 56.2 & -- & -- & -- \\
SegLLM-7B$^{1}$~\citep{segllm_2025} &\citeyear{segllm_2025} & 57.2 & 54.3 & 52.4 & 48.4 \\
UFO-8B$^{3}$~\citep{UFO} &\citeyear{UFO} & -- & -- & 60.0 & -- \\
UniPixel-7B$^{2}$~\citep{UniPixel} &\citeyear{UniPixel} & 60.5 & 58.7 &--&--\\
X-SAM-3.8B$^{5}$~\citep{XSAM} &\citeyear{XSAM} & 56.6 & 32.9 & 57.8 & 41.0\\
HiMTok-8B$^{3}$~\citep{himtok_2025} &\citeyear{himtok_2025} & 60.7 & \underline{67.0} & 60.8 & \underline{66.2}\\
\midrule

\multicolumn{6}{l}{\textit{Textual Localization Readout Method}}\\
SAM4MLLM-7B$^{6}$~\citep{sam4mllm_2024} &\citeyear{sam4mllm_2024} & 46.7 & 48.1 & -- & -- \\
Seg-R1-7B$^{2}$~\citep{seg-r1_2025} &\citeyear{seg-r1_2025} & 58.6 & 41.2 & 56.7 & 53.7 \\
Seg-Zero-7B$^{2}$~\citep{segzero_2025} &\citeyear{segzero_2025} & 62.6 & 62.0 & 57.5 & 52.0\\
VisionReasoner-7B$^{2}$~\citep{VisionReasoner} &\citeyear{VisionReasoner} & \textit{66.3} & -- & \textit{63.6} & --\\
SAM-R1-7B$^{2}$~\citep{SAM-R1} &\citeyear{SAM-R1} & 64.0 & 55.8 & 60.2 & 54.3\\
\midrule

\multicolumn{6}{l}{\textit{Interpretable Alignment Method}}\\
\rowcolor{gray!20}
\modelname{}-7B$^{1}$ & & \textit{66.3} & 65.6 & 61.0 & 63.0 \\
\rowcolor{gray!20}
\modelname{}-7B$^{2}$ & & \underline{66.6} & \textit{66.3} & \underline{64.0} & \textit{64.8} \\

\rowcolor{gray!20}
\modelname{}-13B$^{1}$ & & \textbf{67.4} & \textbf{67.2} & \textbf{64.2} & \textbf{66.5} \\
\bottomrule
\end{tabular}
\end{table}

\minihead{Multi-object Results on gRefCOCO.} We evaluate on the gRefCOCO, a benchmark posing challenges in referring to multiple objects. The comparisons on gRefCOCO are presented in \Cref{tab:main_rst_gref}. Our \modelname{} sets the new state-of-the-art among the 7B and 13B models on most of the splits and metrics, maintaining its competitive edge in multi-object reasoning segmentation tasks. These outcomes highlight the robustness and versatility of \modelname{} in handling more complex segmentation challenges.

\minihead{Zero-shot Results on ReasonSeg.} We evaluate on the ReasonSeg in a zero-shot setting to validate the generalization ability of \modelname{} on complex reasoning segmentation scenarios. From \Cref{tab:main_rst_reasonseg}, our \modelname{} also demonstrates superior results compared to prior state-of-the-art methods. Notably, we find that methods trained with reinforcement learning, such as Seg-Zero, SAM-R1, VisionReasoner, and our \modelname{}, consistently outperform other methods, demonstrating the generalization benefits of reinforcement learning for segmentation models.
\begin{table*}[t]
\centering

% ===== Ablation: training mode =====
\begin{minipage}[htbp]{0.36\textwidth}
\centering
\scriptsize
\setlength{\tabcolsep}{0.55pt}

\captionsetup{type=table,skip=3pt}  % caption 与表格之间改成 3pt（可调 0–6pt）

\captionof{table}{Effect of training mode.}
\label{tab:ablation_trainmode}
\begin{tabular}{c c c c c}
\toprule
\begin{tabular}{@{}c@{}}Reinforcement\\Learning\end{tabular} &
\begin{tabular}{@{}c@{}}Segmentation\\Supervision\end{tabular} &
RefCOCOg & gRefCOCO & ReasonSeg \\
\midrule
\ding{51} & \ding{55} & 65.9 & 63.0 & 40.1 \\
\ding{55} & \ding{51} & 77.9 & 74.0 & 59.3 \\
\rowcolor{gray!10}
\ding{51} & \ding{51} & \textbf{81.3} & \textbf{77.3} & \textbf{66.5} \\
\bottomrule
\end{tabular}
\end{minipage}
\hfill
% ===== Ablation: vision backbone =====
\begin{minipage}[htbp]{0.3\textwidth}
\centering
\scriptsize
\setlength{\tabcolsep}{0.75pt}
\captionsetup{type=table,skip=3pt}  % caption 与表格之间改成 3pt（可调 0–6pt）
\captionof{table}{Effect of vision backbone.}
\label{tab:ablation_backbone}
\begin{tabular}{c c c c c}
\toprule
\begin{tabular}{@{}c@{}}Vision\\Backbone\end{tabular} &
\begin{tabular}{@{}c@{}}Params\\(B)\end{tabular} &
RefCOCOg & gRefCOCO & ReasonSeg \\
\midrule
ViT-B & 0.09 & 78.8  & 75.0 & 58.8 \\
ViT-L & 0.31 & 80.1  & 76.2 & 63.9 \\
\rowcolor{gray!10}
ViT-H & 0.64 & \textbf{81.3} & \textbf{77.3} & \textbf{66.5} \\
\bottomrule
\end{tabular}
\end{minipage}
\hfill
% ===== Ablation: reward function =====
\begin{minipage}[htbp]{0.31\textwidth}
\centering
\scriptsize
\setlength{\tabcolsep}{0.6pt}\captionsetup{type=table,skip=3pt}  % caption 与表格之间改成 3pt（可调 0–6pt）

\captionof{table}{Effect of reward function.}
\label{tab:ablation_reward}
\begin{tabular}{c c c c c}
\toprule
\begin{tabular}{@{}c@{}}Format\\Score\end{tabular} &
\begin{tabular}{@{}c@{}}Segmentation\\Score\end{tabular} &
RefCOCOg & gRefCOCO & ReasonSeg \\
\midrule
0.0 & 1.0 & 79.1 & 75.0 & 63.9 \\
0.5 & 0.5 & 81.0 & 76.8 & 66.2 \\
\rowcolor{gray!10}
0.3 & 0.7 & \textbf{81.3} & \textbf{77.3} & \textbf{66.5} \\
\bottomrule
\end{tabular}
\end{minipage}

\end{table*}

\begin{figure}[h]
\captionsetup{aboveskip=4pt, belowskip=-10pt}
    \centering
    \includegraphics[width=\columnwidth]{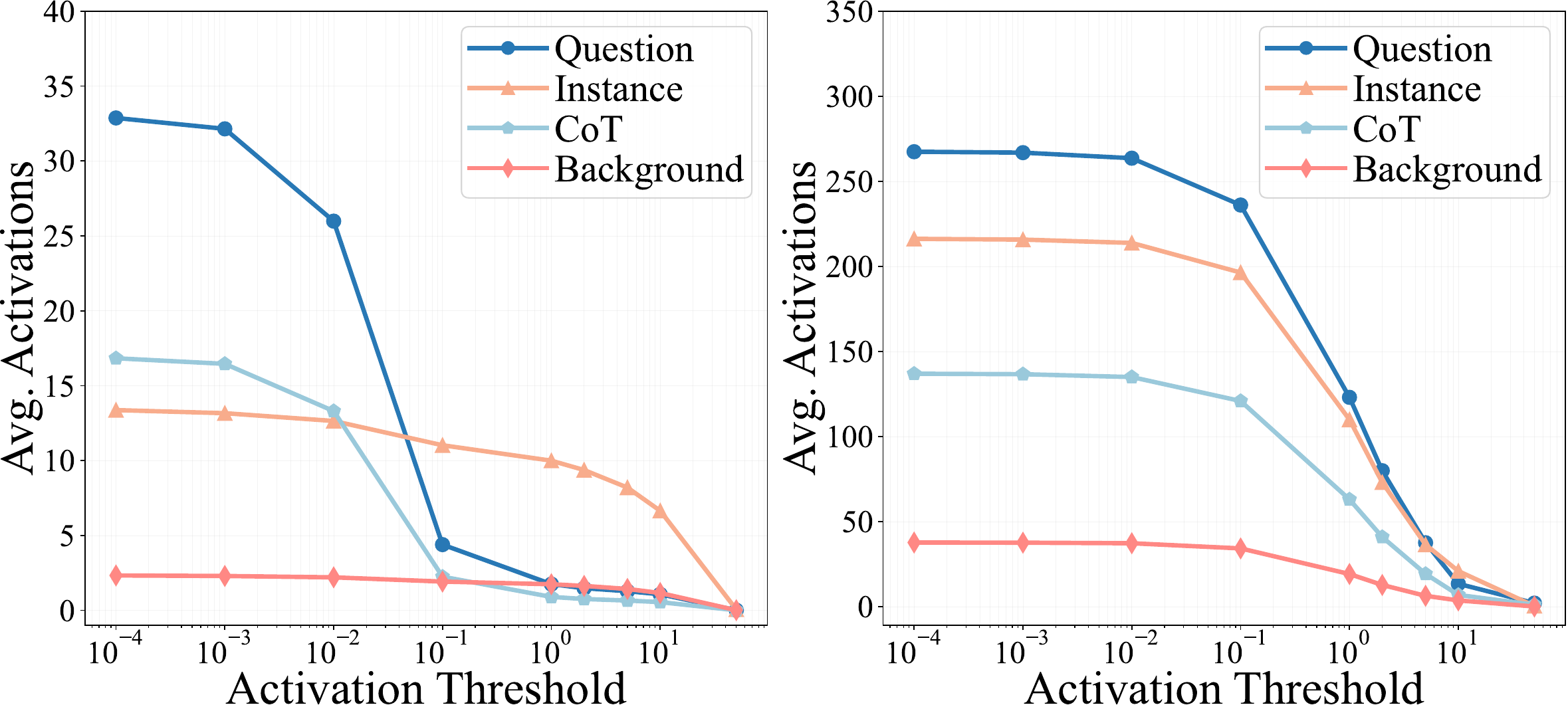}
    \caption{\textbf{Activation patterns.} Average number of active sparse features per token type across activation thresholds. \textbf{Left:} LLaVA-1.5-7B. \textbf{Right:} Qwen2.5-VL-7B.}
    \label{fig:sae_act}
\end{figure}

\subsection{Interpretability Analysis (RQ2)}
\label{subsec:interp_analysis}

\minihead{Interpretable Activation Patterns.}
We analyze the distribution of SAE activations across token types (\Cref{fig:sae_act}).
For a threshold $\tau$, we count the number of active dimensions in the $d_{\text{sae}}=65{,}536$-dimensional space and report the per-token mean for four types: question and CoT tokens from text, and instance and background tokens from image features.
Two patterns emerge.
First, across thresholds, instance and CoT tokens activate the most features. Question tokens are lower but above background, and background is consistently lowest, which indicates semantic selectivity and an interpretable focus on segmented objects and the reasoning process.
Second, most curves are insensitive when $\tau<0.1$, indicating that once a token activates a feature, its activation is typically well above small thresholds rather than marginal noise.
Finally, Qwen2.5-VL-7B shows higher activation counts and larger separations between token types than LLaVA-1.5-7B, consistent with the stronger downstream results reported in \Cref{subsec:overll_performance}.

\begin{figure}[h]
\captionsetup{aboveskip=4pt, belowskip=-10pt}
    \centering
    \includegraphics[width=\columnwidth]{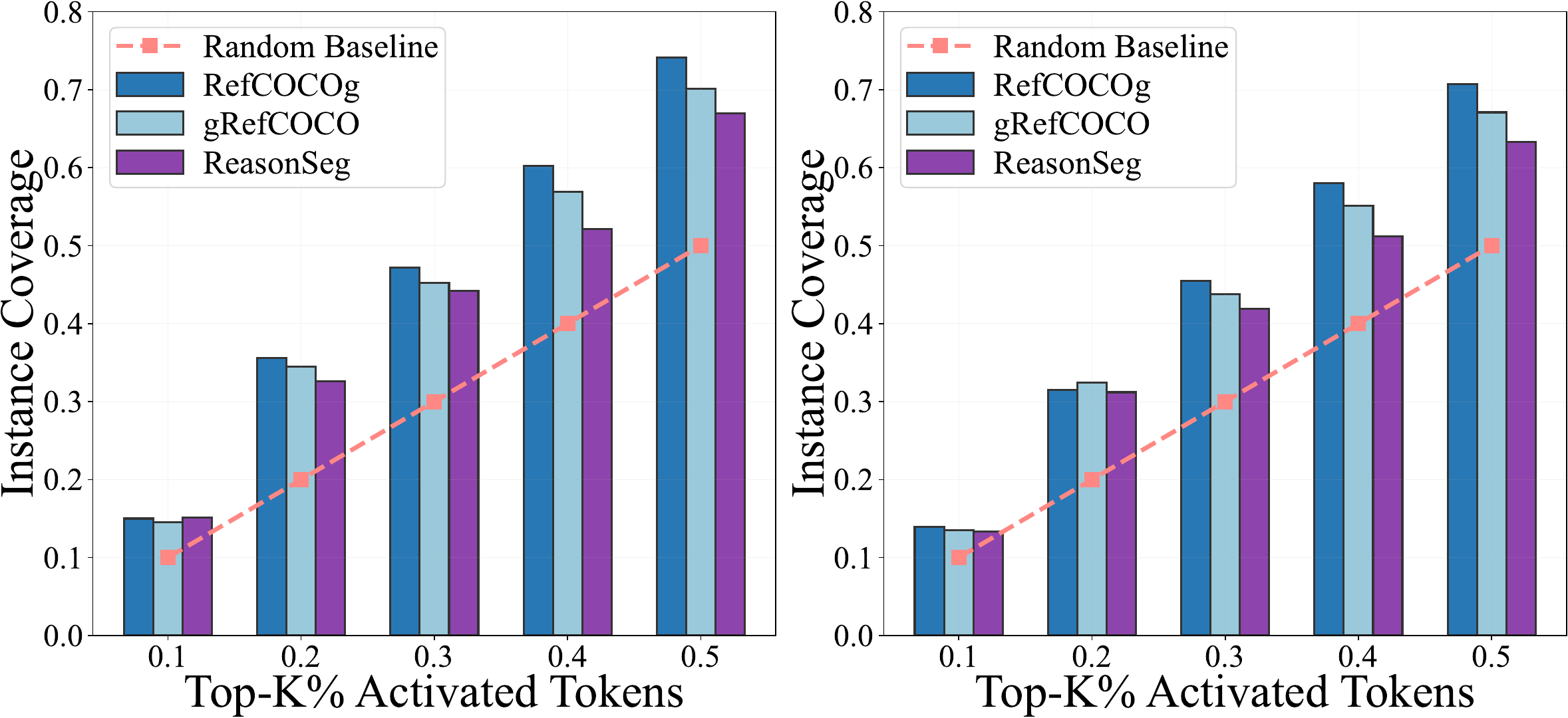}
    % \caption{\textbf{Instance coverage.} The rate instance being covered under varying topK\%. \textbf{Left:} LLaVA-1.5-7B. \textbf{Right:} Qwen-2.5-7B.}
    \caption{\textbf{Instance Coverage Rate vs. Top-K Percentage.} \textbf{Left:} LLaVA-1.5-7B. \textbf{Right:} Qwen2.5-VL-7B.}
    \label{fig:sae_coverage}
\end{figure}

\minihead{Instance Coverage Analysis.}
\Cref{fig:sae_coverage} reports, for each instance, the probability that its pixels are covered by the union of the top-K \% activated tokens, where tokens are ranked by activation magnitude.
A random-K \% baseline (uniformly sampling K \% of tokens) is shown for reference.
Although the SAE is only pretrained and never exposed to segmentation supervision, top-K \% activations yield markedly higher instance coverage than the random baseline, indicating object-sensitive activations aligned with the referred entities.
Moreover, Qwen2.5-VL-7B attains higher coverage than LLaVA-1.5-7B, consistent with the patterns observed in \Cref{fig:sae_act}.
Moreover, we analyze the correlation between modules in Appendix.

\subsection{Ablation Study (RQ3)}
\label{subsec:ablation}
To better demonstrate the impact of different contributing factors, we conduct ablation studies with \modelname{}-13B on RefCOCOg, gRefCOCO, and ReasonSeg. Our investigation centered around the following aspects: the group size of GRPO training, the training mode, the vision backbone of segmentation, and the reward function. These studies provide
insights into the effectiveness and stability of our method.

\begin{figure}[h]
    \captionsetup{aboveskip=4pt, belowskip=-10pt}

    \centering
    \includegraphics[width=\columnwidth]{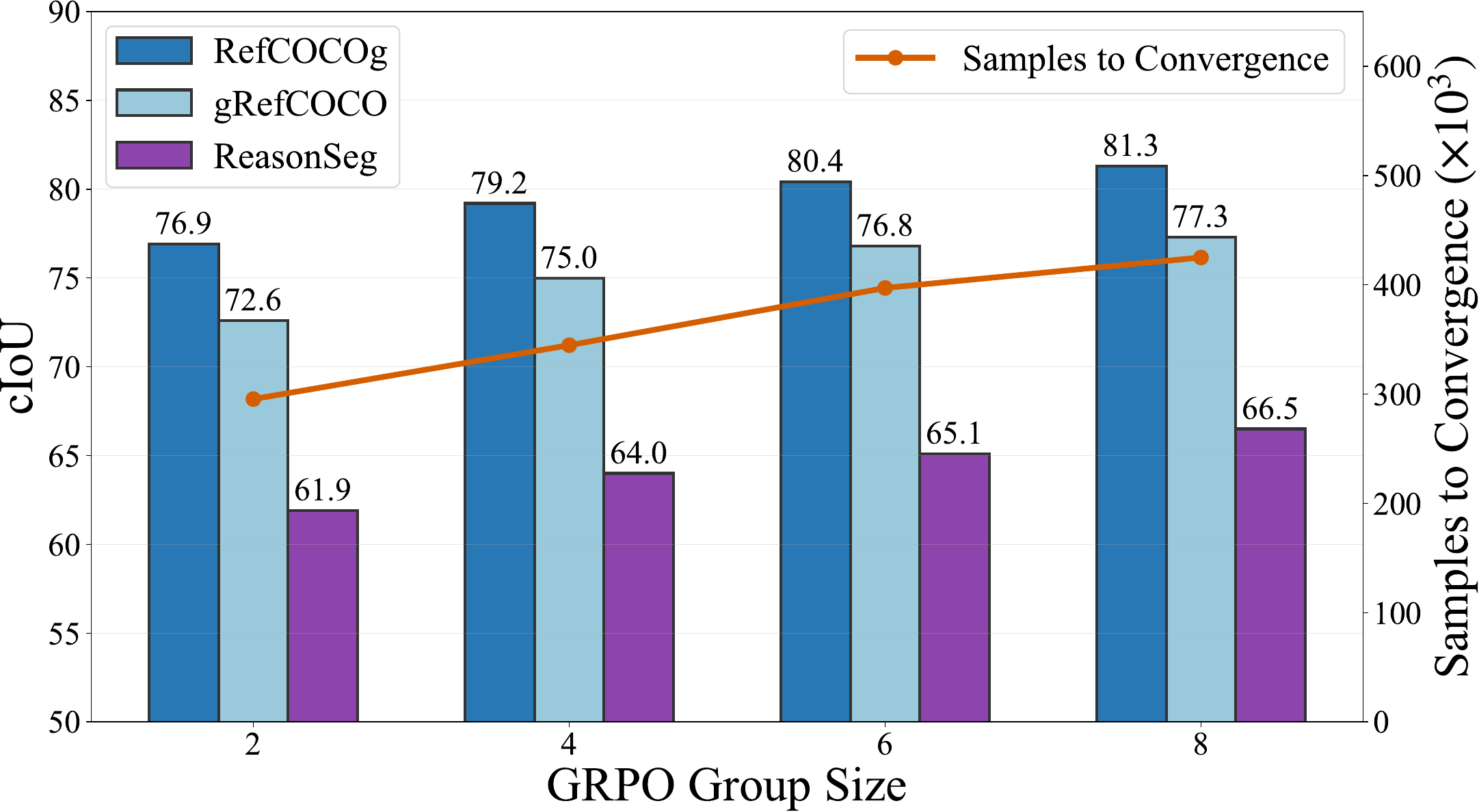}
    \caption{\textbf{Effect of GRPO group size} on performance and sampling cost. We report cIoU vs. group size $G$ on RefCOCOg, gRefCOCO, and ReasonSeg, and plot the total number of sampled responses required for convergence.}
    \label{fig:grpo_n}
\end{figure}

\minihead{GRPO Group Size.} The group size refers to the number of answers sampled for each question during rollout and is the important parameters of GRPO. We conduct experiments on different group sizes, as shown in \Cref{fig:grpo_n}. We can see that increasing the group size can achieve effective performance improvement. The advantage of a larger group size is that it allows the model to see a larger sample space, and the difference between positive and negative samples within the group is more pronounced. In addition, we also count the total number of rollout samples when the training loss fluctuates less than 5\% within a certain step (i.e.\ Samples to Convergence) in \Cref{fig:grpo_n}. We find that, under different group sizes, the difference in the total number of rollout samples during GRPO training convergence is small, but larger group sizes can achieve better performance.

\minihead{Training Mode.} We compare the effect of different training modes (i.e.\ training objective function), including reinforcement learning, segmentation supervision, and a combined objective, for \modelname{}-13B. As shown in \Cref{tab:ablation_trainmode}, the combined objective achieves the best performance. The reason for this phenomenon is that reinforcement learning can enhance reasoning, while supervised signals can enhance mask generation. They are more effective together for complex reasoning segmentation.

\minihead{Vision Backbone.} We verify the impact of the size of visual backbone on the segmentation performance of \modelname{}-13B. The experimental results are presented in \Cref{tab:ablation_backbone}. From the experimental results, it can be found that larger visual backbones have better segmentation performance. The larger visual backbone has stronger representation capacity and can provide more information for decoding segmentation results.

\minihead{Reward Function.} We also evaluate alternative reward designs. \Cref{tab:ablation_reward} compares combinations of the format score and the segmentation score. The segmentation score contributes more to segmentation performance. This advantage stems from the segmentation score's direct correlation with the quality of segmentation results, while the format score only regulates the behavior of the model.

\section{Conclusion}
\label{sec:conclusion}

In this work, we introduced \modelname{}, a novel framework that successfully bridges the interpretability gap in reasoning segmentation. Our core innovation is an SAE that forges an explicit ``white-box" alignment between CoT reasoning and mask prediction. The SAE maps both CoT and visual tokens into a shared sparse concept space, which is then grounded into a multi-slot heatmap to guide the final segmentation. This end-to-end framework, jointly trained with GRPO and segmentation supervision, bypasses the opacity of latent queries and the ambiguity of textual readouts.
\modelname{} achieves state-of-the-art performance across five challenging benchmarks. Our quantitative and visual analyses confirm that this strong performance is directly linked to the quality of the learned sparse concepts, which provide an inspectable pathway from reasoning to perception.

\section*{Acknowledgments}
We sincerely thank the anonymous reviewers and chairs for their efforts and suggestions, greatly helping us improve the manuscript. This work is supported in part by the National Natural Science Foundation of China under grants 62536003 and 624B2088, and in part by the project of Peng Cheng Laboratory (PCL2025A14).

{\small
\bibliographystyle{ieeenat_fullname}
\bibliography{main}
}

\clearpage
\setcounter{page}{1}
\maketitlesupplementary
\appendix
\section{Implementation Details}
\subsection{Pipeline Implementation}
\minihead{Parallelism and Optimization.} Our implementation leverages the VERL~\citep{verl_2025} framework, adapted for multi-modal GRPO~
\citep{grpo_2024} training. To manage memory efficiency, we utilize Fully Sharded Data Parallel (FSDP)~\citep{fsdp_2023} to partition the MLLM policy parameters across devices. The lightweight segmentation modules, comprising the query head, Q–V attention~\citep{attention_2017}, and mask decoder, are kept unsharded to avoid unnecessary communication overhead. For inference acceleration, the vLLM~\citep{vllm_2023} backend employs tensor parallelism across attention heads. Additionally, image features are precomputed offline, removing the frozen vision backbone from the active training graph.
\minihead{Distributed Worker Architecture.} To reduce computational overhead, we precompute image features offline, thereby excluding the frozen vision backbone from the training loop.
Our framework orchestrates three distinct types of FSDP workers:
(i) The \textbf{actor} contains all trainable modules (including the MLLM and segmentation components) and is responsible for gradient computation and parameter updates.
(ii) The \textbf{rollout worker} executes the MLLM in inference mode, processing image and text inputs to generate responses trajectories via next-token prediction.
(iii) The frozen \textbf{reference worker} maintains a copy of the reference policy model to compute the KL divergence term for $\Lgrpo$. Additionally, it utilizes the segmentation modules to decode masks required for calculating mask-based reward scores and group advantages. \par

\minihead{Training Workflow.} For each annotated sample, the rollout worker first generates a group of $G$ responses using the current policy model via next-token predict, caching both the output tokens and their log probabilities. Subsequently, the frozen reference worker performs a forward pass (without gradients) on the same inputs. This step computes  the reference log probabilities for the sampled responses and decode masks for the mask-based reward. For each response and its mask signal, we compute a scalar reward and convert rewards to group advantages. Finally, the actor worker performs a forward pass to obtain the current policy log probabilities for the sampled responses and the predicted segmentation mask. The total loss is composed of the GRPO objective (derived from actor log probabilities, cached rollout log probabilities, reference log probabilities, and advantages) and the segmentation objective (derived from the predicted and ground truth mask. The two objectives are summed and jointly optimized in a single backward pass to update all trainable parameters.

\begin{figure}[ht]
\captionsetup{aboveskip=6pt, belowskip=-8pt}
    \centering
    \includegraphics[width=\columnwidth]{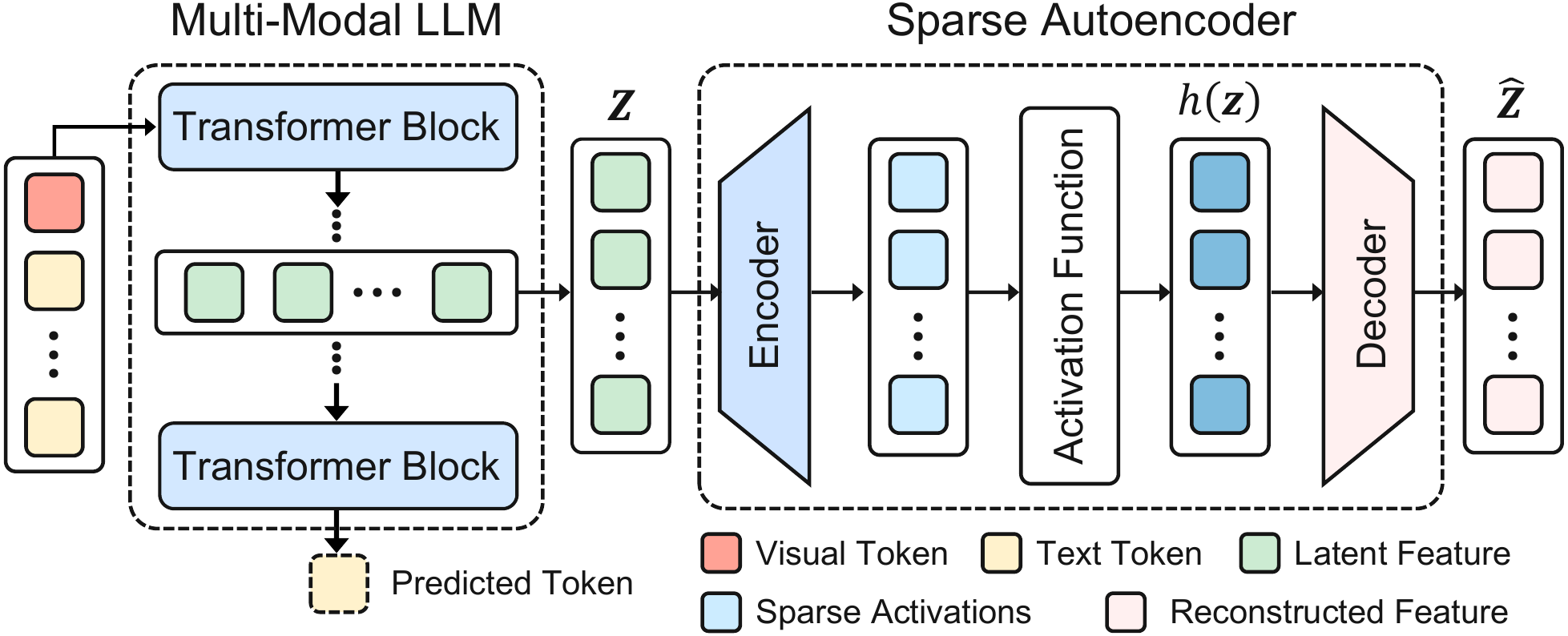}
    \caption{The architecture of Sparse Autoencoder (SAE). }
    \label{fig:sae_architecture}
\end{figure}

\minihead{Sparse Autoencoder (SAE).} We adopt the architecture of SAE from the methodology proposed in SAE-V \cite{sae-v_2025}, which comprises an encoder and a decoder, as shown in~\Cref{fig:sae_architecture}. The input feature of SAE is the hidden layer state of MLLM. We maintain consistent experimental configurations across LLaVA-1.5-7B~\citep{llava1.5_2024}, LLaVA-1.5-13B and Qwen2.5-VL-7B~\citep{qwen2.5_vl}, with the exception of the specific Transformer decoder layer selected for feature extraction. Specifically, we extract features from the 16-th layer for both LLaVA-1.5-7B (32 layers total) and LLaVA-1.5-13B (40 layers total), and the 13-th layer for Qwen2.5-VL-7B (28 layers total). Following the standard practice in SAE, we target these intermediate layers to strike a balance between modality fusion and semantic abstraction. Middle layers are empirically found to encode rich mixed vision-language information, whereas deeper layers tend to be dominated by text-generation patterns (i.e., next-token prediction dynamics). SAE training is conducted before the reasoning segmentation task, during which its parameters are frozen.

\subsection{Training Configuration}
\minihead{Data Preprocessing.} We train on the RefCOCO~\citep{refcoco_withplus_2014, refcocog_2016} series and gRefCOCO~\citep{ReLA_2023} datasets. The maximum prompt length and response length are set to 1,400 tokens and 2,000 tokens, respectively. For Qwen2.5-VL-7B, the input image pixel count range from 3,136 to 705,600. Image exceeding this range are resized accordingly. We initialize the vision branch using SAM ViT-H.

\minihead{Model Optimization.} We optimize the policy model parameter using AdamW with a weight decay of \(0.01\) and \((\beta_1,\beta_2)=(0.9,0.999)\). We employ differential learning rates across model components: the base learning rate for the MLLM backbone is set to \(2\times 10^{-6}\), with multipliers of $25\times$ for the query codebook, $80 \times$ for the slot mapper, and $10\times$ for the mask decoder. Gradient clipping is applied with a maximum norm of 1.0. The schedule is one cycle with a final division factor of 6.7 and no warmup. Training spans 24,252 steps with gradient checkpointing enabled. The global batch size is 128, comprising 16 image-text pairs with a group size of 8 for GRPO. We utilize bfloat16 precision for model parameters and fp32 for reductions and buffers. For SAE training, we set the learning rate, batch size, and total training epochs to \(5\times 10^{-5}\), 64, 10, respectively. Additionally, We employ a warmup phrase during the initial stage of training.

\minihead{GRPO Settings.} We configure GRPO with a group size of 8, a clip ratio of 0.2, and a fixed KL penalty coefficient 0.2. For the policy model, the per-device micro-batch is set to 2 for updates and 8 for experience collection. The policy model is trained using Fully Shared Data Parallel (FSDP) with optimizer state offloading. For the rollout service, we employ a vLLM backend with a tensor parallelism size of 4 for sampling. Inference is executed in bfloat16, supporting a maximum of 64 concurrent sequences and a batch size of 17,408 batch tokens. Chunked prefill is enabled and each sample consists of a single image.

\section{Group Relative Policy Optimization}
Recently, Reinforcement Learning (RL) has been applied to enhance the reasoning capability of LLMs. By formulating reasoning token generation as a Markov Decision Process, models are trained to maximize the expected return of reasoning paths, guiding optimization toward more structured and coherent reasoning trajectories. Proximal Policy Optimization (PPO)~\citep{ppo} and its variants, Group Relative Policy Optimization (GRPO)~\citep{grpo_2024, deepseek-r1_2025}, are among the most widely adopted algorithms for this purpose.

\subsection{PPO Preliminaries}
PPO is an actor-critic RL algorithm that optimizes LLMs by maximizing the following surrogate objective:
\begin{equation}
\resizebox{1\hsize}{!}{$
\begin{aligned}
\Lppo = &\mathbb{E}\left[q \sim P(Q), o \sim \pi_{{\bm \theta}_{\text{old}}}(O|q)\right]\\& \frac{1}{|o|} \sum_{t=1}^{ |o|} \min \left[ \frac{\pi_{{\bm \theta}}(o_t|q, {\bm o}_{<t})}{\pi_{{{\bm \theta}}_{\text{old}}}(o_t|q, {\bm o}_{<t})} A_t, \text{clip} \left( \frac{\pi_{{\bm \theta}}(o_t|q, {\bm o}_{<t})}{\pi_{{{\bm \theta}}_{\text{old}}}(o_t|q, {\bm o}_{<t})}, 1 - \varepsilon, 1 + \varepsilon \right) A_t \right]
\end{aligned}$}
\end{equation}\noindent\ignorespaces
where~$\pi_{\bm \theta}$ and~$\pi_{{\bm \theta}_{old}}$ denote the current and old policy models, respectively. Here,~$q$ and~$o$ represents questions and outputs sampled from the dataset and the old policy~$\pi_{{\bm \theta}_{old}}$. The~$\varepsilon$ is a clipping-related hyper-parameter introduced in PPO for stabilizing training. The advantage~$A_t$ is computed using Generalized Advantage Estimation (GAE), based on the reward~$\{r_{\geq t}\}$ and a learned value function~$V_{\psi}$. The value function~$V_{\psi}$ is typically approximated by a learnable critic model of the same scale as the policy model. Furthermore, a per-token KL penalty from a reference model is added to the reward at each step to mitigate over-optimization of the reward model, which denoted as:
\begin{equation}
r_t = r_{\varphi}(q, {\bm o}_{\leq t}) - \beta \log \frac{\pi_{\bm \theta}(o_t | q, {\bm o}_{<t})}{\pi_{\text{ref}}(o_t | q, {\bm o}_{<t})}
\end{equation}\noindent\ignorespaces
where~$r_{\varphi}$ represents the reward model,~$\pi_{\text{ref}}$ denotes the reference model (typically the initial policy model), and~$\beta$ is the coefficient of the KL penalty. In PPO, both the policy and critic models must be trained simultaneously, which imposes significant computational demands when model parameters or token counts are large. GRPO has been proposed to reduce the resource consumption associated with PPO while maintaining training stability.

\subsection{GRPO Objective}
GRPO eliminates the need for a separate value function approximator as in PPO. Instead, it estimates the advantage using the average reward of multiple outputs sampled for the same question. Specifically, for each question~$q$, GRPO samples a group of outputs~$\{o_1, o_2, \cdots, o_{\scriptscriptstyle G}\}$ from the old policy~$\pi_{{\bm \theta}_{\text{old}}}$ and optimizes the policy model by maximizing the following objective:
\begin{equation}
\resizebox{1.0\linewidth}{!}{$
\begin{aligned}
\Lgrpo = &\mathbb{E}\left[q \sim P(Q), \{o_i\}_{i=1}^G \sim \pi_{{\bm\theta}_{\text{old}}}(O|q)\right]\\ &
\frac{1}{G} \sum_{i=1}^G \min \left[ \frac{\pi_{\bm \theta}(o_{i}|q)}{\pi_{{\bm\theta}_{\text{old}}}(o_{i}|q)} \hat{A}_{i}, \text{clip} \left( \frac{\pi_{\bm\theta}(o_{i}|q)}{\pi_{{\bm\theta}_{\text{old}}}(o_{i}|q)}, 1 - \varepsilon, 1 + \varepsilon \right) \hat{A}_{i} \right] - \beta \mathbb{D}_{\text{\scriptsize \textsc{kl}}} \left[ \pi_{\bm\theta} \| \pi_{\text{ref}} \right]
\end{aligned}$}
\end{equation}\noindent\ignorespaces
where~$\varepsilon$ and~$\beta$ are hyper-parameters, and~$\hat{A}_{i}$ represents the advantage computed using the relative rewards of outputs within each group. For each question~$q$, a group of outputs~$\{o_1, o_2, \cdots, o_G\}$ is sampled from the old policy model~$\pi_{{\bm\theta}_{old}}$. A reward model assigns scores to these outputs, yielding~$G$ corresponding rewards~$\{r_1, r_2, \cdots, r_G\}$. The advantages ~$\hat{A}_{i}$ of the tokens in the output is defined as the normalized reward:~$\hat{A}_{i} = \tilde{r}_i = \frac{r_i - \text{mean}(\bm{r})}{\text{std}(\bm{r})}$. Additionally, GRPO incorporates the KL divergence between the trained policy and the reference policy directly to the loss function, avoiding complicating the calculation of~$\hat{A}_{i}$. The KL divergence is estimated by the following unbiased estimator:
\begin{equation}
\mathbb{D}_{\text{\scriptsize \textsc{kl}}} \left[
,\pi_{\bm\theta} \,\|\, \pi_{\text{ref}} \,\right] = \frac{\pi_{\text{ref}}(o_{i}|q)}{\pi_{\bm\theta}(o_{i}|q)} - \log \frac{\pi_{\text{ref}}(o_{i}|q)}{\pi_{\bm\theta}(o_{i}|q)} - 1
\end{equation}\noindent\ignorespaces

\subsection{Reward Function Design}
\minihead{Format Score.}
We score outputs in $\{0.0, 0.9, 1.0\}$. A response is valid only if it contains exactly one \texttt{<think>} ... \texttt{</think>} block with non-empty content and exactly one special token for reference position; the special token must appear after \texttt{</think>}. Any violation yields 0.0. For valid outputs, the base is 1.0 and is downgraded to 0.9 if the \texttt{<think>} content is overly long (more than 2048 characters), or if there is any non-whitespace text before \texttt{<think>} or after the special token. Thus the five canonical cases are: invalid (0.0); valid and clean (1.0); valid but long \texttt{<think>} (0.9); valid but extra text before \texttt{<think>} (0.9); valid but extra text after the special token (0.9).

\minihead{Single-object.}
We use a scalar reward that combines a format score with a mask score. For the single-object setting, no assignment is required: the mask score is computed directly from the agreement between the predicted probability mask and the ground-truth mask using a soft IoU overlap in $[0,1]$. This measures how well the prediction localizes the referent without introducing any auxiliary terms or matching.

\minihead{Multi-object.}
For multiple instances, the model outputs K slot masks with confidences. We retain predictions above a fixed confidence threshold and compare them to the G ground truth masks via pairwise soft overlaps. A Hungarian assignment is then solved to maximize total overlap, and the mask score is the mean overlap over the matched pairs. If one side is empty and the other is not, the score is set to 0; if both are empty, it is set to 1.0. The final reward still combines the format score and this mask score, encouraging valid reasoning traces and precise one-to-one coverage in the multi-object case.

\section{Additional Analyses}
\subsection{Correlation Analyses}
To verify the effectiveness of the designed concept representation and concentration token in guiding the segmentation process, we examine two key correlations: (1) the correlation between the SAE feature and segmentation during training, and (2) the correlation between the heatmap and segmentation results during inference.
\begin{figure}[h]
    \centering
    \includegraphics[width=\columnwidth]{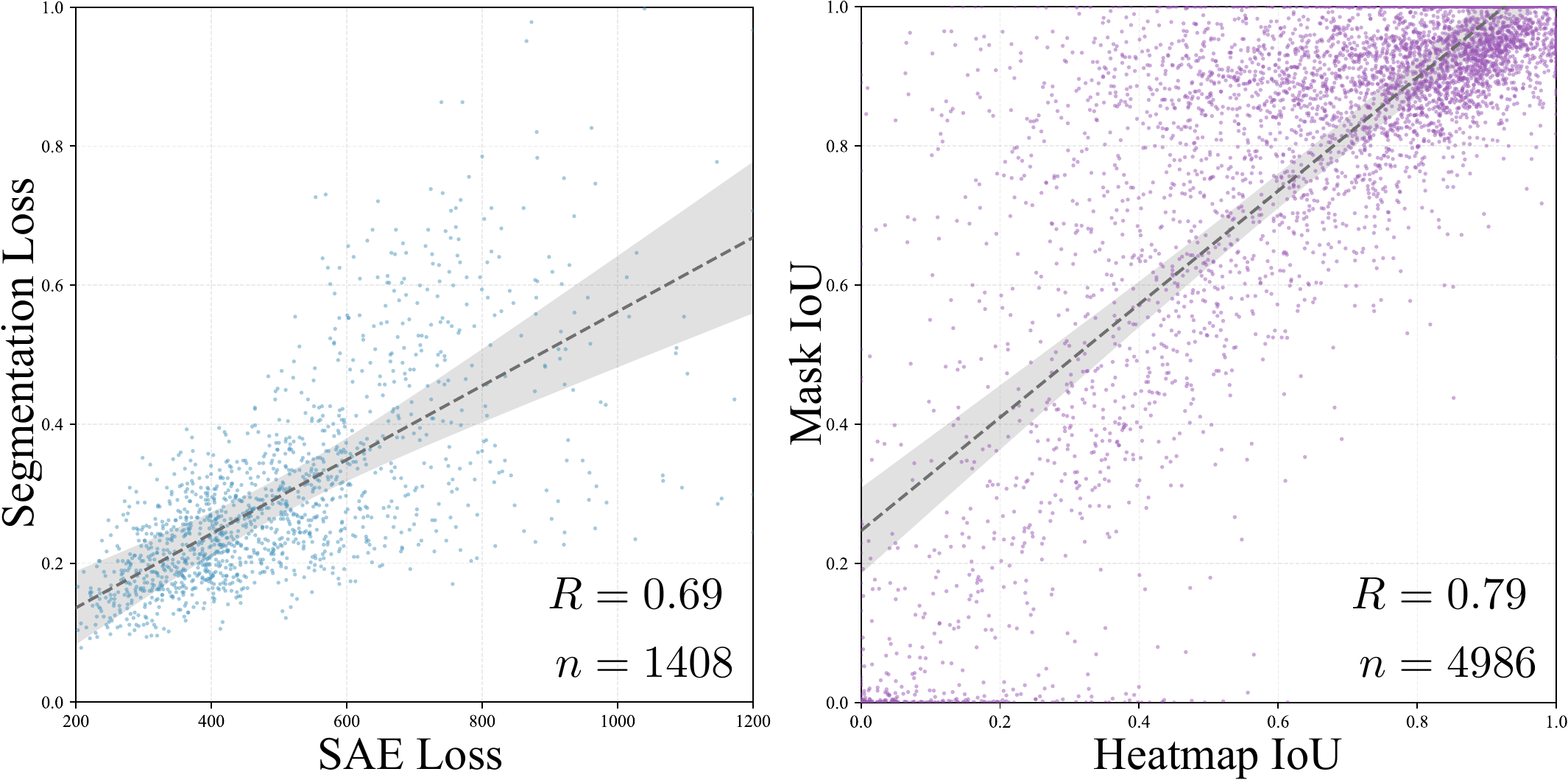}
    \caption{\textbf{Correlation analysis.} Left: SAE features vs. segmentation during training. Right: heatmap vs. segmentation during inference. Results for Qwen2.5-VL-7B on gRefCOCO. Ordinary least squares (OLS) regression lines and mean confidence bands are overlaid.}
    \label{fig:correlation_4_3}
\end{figure}
In \Cref{fig:correlation_4_3} (left), each data point denotes a training batch.
The x-axis displays the SAE reconstruction error (MSE) while the y-axis represents the training Dice loss of the predicted masks.
Note that the SAE is pretrained and frozen within the segmentation pipeline.
Consequently, MSE measures SAE reconstruction quality, whereas Dice loss reflects mask accuracy.
The scatter plot reveals a distinct trend, indicating a strong association between SAE reconstruction quality and segmentation performance.
In \Cref{fig:correlation_4_3} (right), each point corresponds to a test instance.
The x-axis represents the cIoU between the heatmap $\heatmap$ and the ground truth $\Mgt$, while the y-axis shows the cIoU between the predicted mask $\hat{\bm{M}}$ and $\Mgt$.
We observe a significant positive relationship. These finds demonstrate that both SAE and heatmap quality are positively associated with final mask performance, providing evidence that the interpretable alignment contributes predictive signal.

\subsection{Additional Visualizations}
\begin{figure*}[t]
    \centering
    \includegraphics[width=\linewidth]{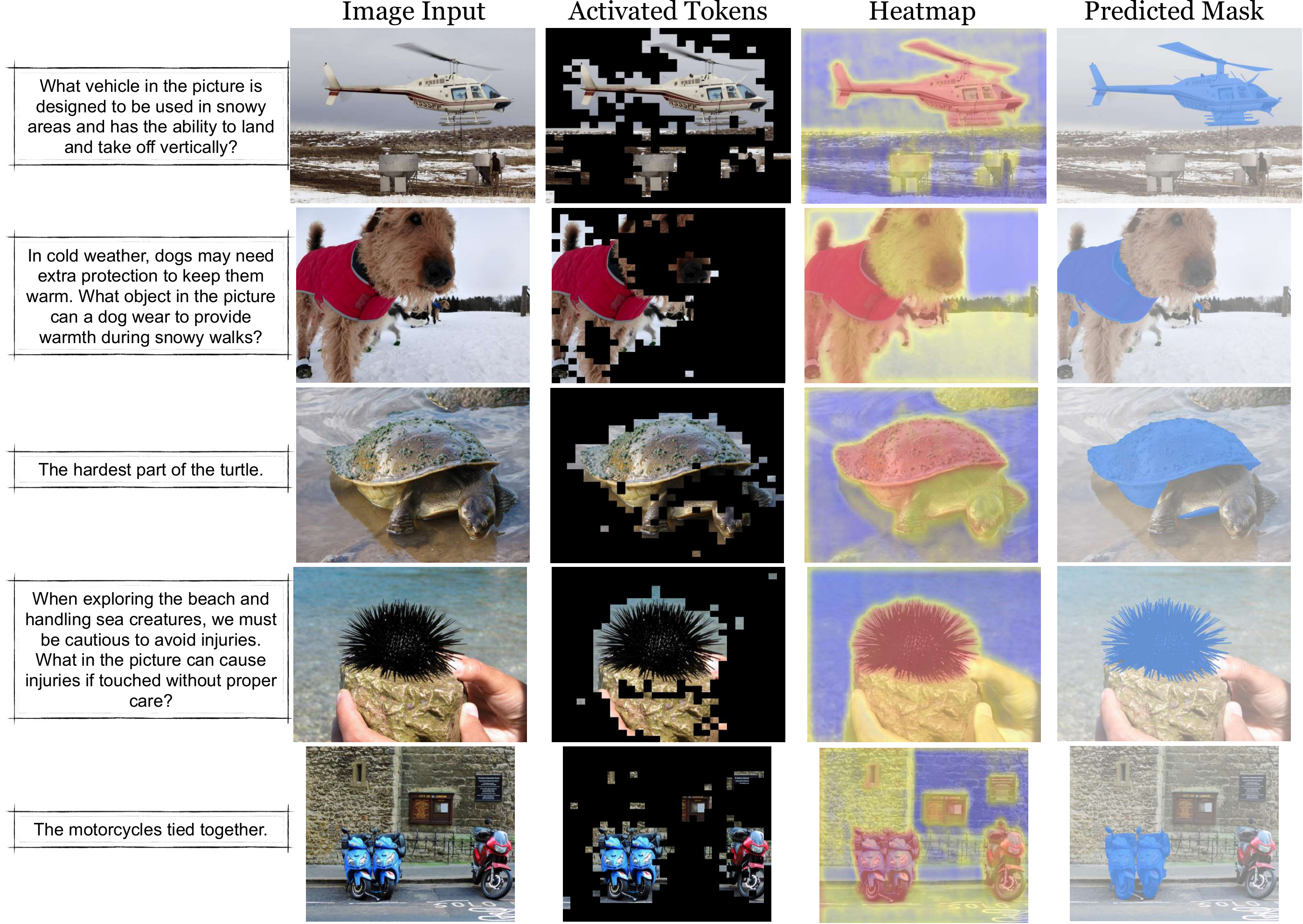}
    \caption{\textbf{Additional Visualizations.} From left to right: instruction, image input, activated tokens, heatmap, and predicted mask.}
    \label{fig:appendix_vis}
\end{figure*}

As shown in \Cref{fig:appendix_vis}, we visualize cases from the ReasonSeg test split. Tokens in regions that match the instruction’s semantics show higher responses in the heatmap, and the map peaks on the referred instance, providing a gook prior that helps the decoder recover clean boundaries. In addition, SAE activations over image tokens are tightly linked to the question: high-activation tokens concentrate on foreground areas and correlate with the heatmap intensity. This behavior is consistent with the quantitative trend in \Cref{fig:correlation_4_3}, where SAE activations for question and instance tokens exceed those for background and instruct the heatmap signal.

\end{document}